\documentclass{article} 
\usepackage[T1]{fontenc}
\usepackage[preprint]{iclr2027_conference}
\usepackage{times}

\usepackage{amsmath,amssymb}
\usepackage{booktabs,array}
\usepackage{multirow}
\usepackage{graphicx}
\usepackage{xcolor}
\usepackage{float}
\usepackage{algorithm}
\usepackage{algpseudocode}
\usepackage{url}
\usepackage{tikz}
\usetikzlibrary{arrows.meta,positioning,shapes.geometric,calc}
\usepackage{pgfplots}
\pgfplotsset{compat=1.18}
\usepackage[hidelinks]{hyperref}
\usepackage{listings}
\usepackage{enumitem}
\newenvironment{myitemize}
  {\begin{itemize}[
    topsep=2pt,
    itemsep=1pt,
    parsep=0pt,
    partopsep=0pt,
    leftmargin=*
  ]}
  {\end{itemize}}
\lstdefinestyle{prog}{language=Python,basicstyle=\ttfamily\fontsize{5.9pt}{6.9pt}\selectfont,columns=fullflexible,keepspaces=true,
  breaklines=false,showstringspaces=false,commentstyle=\color{black!55},keywordstyle=\bfseries,stringstyle=\color{black},
  frame=single,framerule=0.3pt,rulecolor=\color{black!40},xleftmargin=2pt,xrightmargin=2pt,aboveskip=1pt,belowskip=2pt,
  moredelim=**[is][\bfseries\color{orange!85!black}]{<<}{>>},moredelim=**[is][\bfseries\color{blue!65!black}]{!!}{!!}}

\newcommand{\vetr}{VETR}

\newcommand{\nclip}{N\text{-}CLIP}
\newcommand{\PL}{PL}
\newcommand{\best}[1]{\textbf{#1}}
\newcommand{\ci}[2]{{\footnotesize[$#1$,\,$#2$]}}
\title{VETR: Execution-Grounded Self-Improving Agent for Symbolic Graphics Reasoning}
\author{\textbf{Yingbo Ma\textsuperscript{1,*}, Kuangshi Ai\textsuperscript{2,*}, Gordon Dai\textsuperscript{1,*}, Chenglong Yu\textsuperscript{1}, Chaoli Wang\textsuperscript{2}, Hua Shen\textsuperscript{1}}\\[3pt]
{\normalfont\small \textsuperscript{1}~New York University}\\
{\normalfont\small \textsuperscript{2}~University of Notre Dame}}
\hypersetup{
  pdftitle={VETR: Execution-Grounded Self-Improving Agent for Symbolic Graphics Reasoning},
  pdfauthor={Yingbo Ma, Kuangshi Ai, Gordon Dai, Chenglong Yu, Chaoli Wang, Hua Shen}
}

\begin{document}
\maketitle
\begingroup
\renewcommand{\thefootnote}{\fnsymbol{footnote}}
\footnotetext[1]{Equal contribution.}
\endgroup
\raggedbottom
\begin{abstract}
Visual agents for symbolic graphics reasoning depend on executable world representations, yet continued improvement remains bottlenecked by ambiguous supervision from final renders. Recent agentic systems refine scenes through repeated interaction, but they lack mechanisms to recursively extract actionable evidence from execution traces, overlooking which proposed symbolic edits survived, disappeared, or interfered during program execution. To address this, we propose \vetr{} (\underline{V}erified \underline{E}xecution \underline{T}rajectories for \underline{R}efinement), a framework that enables \emph{agentic recursive self-improvement} over executable world representations for symbolic graphics reasoning. Within each task, \vetr{} retains aligned program/render records with bounded runtime observations and introduces an \emph{execution refiner} that compiles observed writes and state changes into constrained alternative programs. The refiner validates their endpoints and preservation conditions against captured runtime state; the ordinary visual verifier then assesses goal relevance. Repair construction makes no additional language-model calls, but can require additional executions. Evaluations on BlenderGym (280 tasks) and BlenderBench (27 released tasks) with GPT-5.6-luna and Qwen3.8-27B show that \vetr{} consistently outperforms existing refinement strategies, including BlenderAlchemy and VIGA. Controlled ablations evaluate the contributions of execution refinement and role-asymmetric evidence organization under fixed generator-interaction caps.
\end{abstract}

\section{Introduction}
\label{sec:intro}

Large Language Model (LLM) agents have made remarkable strides in code generation, visual reasoning, and interactive editing by equipping LLMs with tools, planning, and self-refinement capabilities~\citep{madaan2023selfrefine,shinn2023reflexion,gou2024critic}. Building on this progress, recent systems for symbolic graphics reasoning formulate scene understanding and editing as an iterative process where a generator proposes program modifications and a verifier evaluates the rendered output~\citep{yin2026viga,kuang2026vulcan}. These systems demonstrate that agentic interaction improves over one-shot generation and that executable programs can serve as world representations specifying geometry, materials, lighting, and spatial relationships~\citep{wang2026codeasworld}. However, final-render feedback alone does not expose edit causality: an image comparison need not reveal which proposed symbolic edits survived execution, which were silently overwritten, or which interfered with shared scene resources.

This motivates an execution-evidence interface alongside visual feedback~\citep{yin2026viga,kuang2026vulcan}. Additional iterations can supply new proposals, but do not directly expose effects omitted from the feedback interface. Logging alone also leaves open how observed values should become alternative programs~\citep{wang2026codeasworld}. We address this with supported update operators and explicit runtime preservation checks. Observed writes supply hypotheses about alternative executions, while visual verification remains responsible for assessing those alternatives against the task goal.

\begin{figure}[h]
  \centering
  \includegraphics[width=\linewidth]{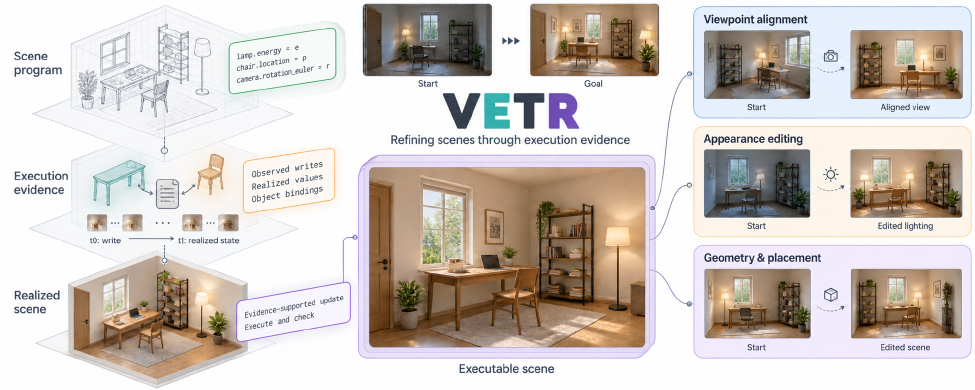}
  \caption{\textbf{Conceptual overview of \vetr{}.}
 VETR refines scenes through execution-grounded evidence, bridging a scene program, its execution trace, and the realized 3D scene. Given a start and goal image, the system performs iterative updates guided by observed writes and realized bindings. This evidence-supported refinement targets viewpoint alignment, appearance (lighting/material) editing, and geometry and placement adjustments. Runtime checks test supported effects, while visual feedback guides subsequent proposals toward the goal.}
  \label{fig:vetr-overview}
\end{figure}

To address this, we propose \textbf{VETR} (\underline{V}erified \underline{E}xecution \underline{T}rajectories for \underline{R}efinement), a framework for execution-grounded, within-task program refinement. VETR retains aligned executable records and uses a private execution refiner to construct alternatives from observed writes and state transitions. The refiner compiles supported numerical changes and compatible resource bindings, then checks their endpoints and preservation conditions by fresh execution, without additional language-model calls. It delivers a runtime-valid alternative when available, otherwise retaining the raw successful result or the preceding delivered record after failure. The ordinary visual verifier assesses the delivered result, and the generator conditions its next proposal on the available execution outcomes and feedback. This is the recursive program-revision loop depicted in Figure~\ref{fig:vetr-overview}; Algorithms~\ref{alg:executionrefiner} and~\ref{alg:wide} specify its refiner and host loop.

We evaluate VETR on BlenderGym (280 editing tasks across five families) and BlenderBench (27 released tasks at three difficulty levels) with GPT-5.6-luna and Qwen3.8-27B. VETR is the strongest complete system on both benchmarks and both backbones, and paired comparisons confirm gains are not artifacts of completion rates. Ablations isolate the improvement to the execution refiner (disabling it reduces six-cell BlenderBench improvement from $+54.45\%$ to $+35.33\%$ on GPT and $+56.17\%$ to $+30.09\%$ on Qwen), and to role asymmetry and execution history (with GPT on BlenderGym, removing role asymmetry or limiting history raises $\nclip$/$\PL$ errors by 12.7\%/23.7\% and 48.5\%/79.0\%, respectively). These results support contributions from execution refinement and evidence organization within the evaluated agent loop; fixed generator-interaction caps do not imply equal total execution cost.

Our contributions are summarized as follows:
\begin{myitemize}
    \item \textbf{Executable world representation for symbolic graphics reasoning.} We build on scene representation work to establish executable programs as the foundation for iterative scene editing, enabling direct manipulation of geometry, materials, lighting, and spatial relationships.
    \item \textbf{Agentic recursive self-improvement with an execution refiner.} We formulate scene refinement as a propose--execute--observe--verify cycle and design a refiner that compiles execution-trace evidence into bounded program updates under preservation constraints, without additional LLM inference~\citep{madaan2023selfrefine,shinn2023reflexion,gou2024critic}.
    \item \textbf{Empirical validation on graphics editing benchmarks.} We evaluate on BlenderGym and BlenderBench across two backbones, showing consistent gains over VIGA~\citep{yin2026viga} and BlenderAlchemy, with controlled ablations evaluating execution refinement and context organization within the host loop.
\end{myitemize}

\begin{figure}[t]
\centering
\includegraphics[width=\linewidth]{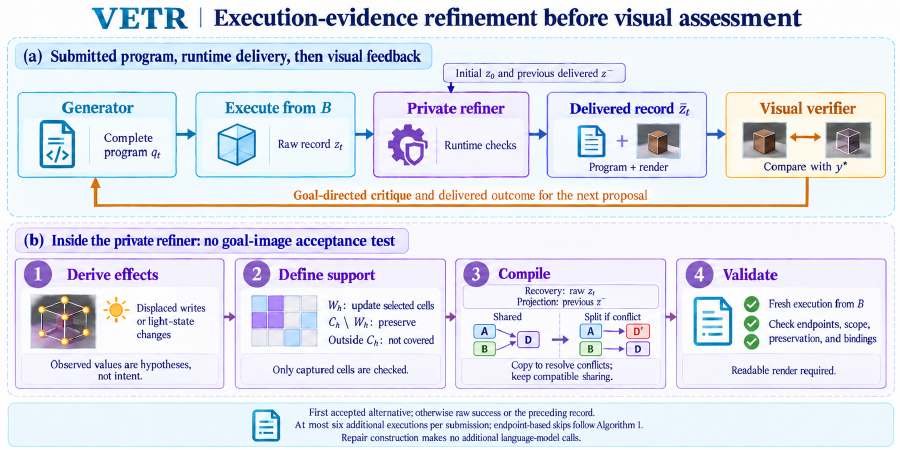}
\caption{\textbf{Execution-evidence refinement before visual assessment.} Each complete program executes from $B$. The private refiner constructs alternatives, compiles supported changes and compatible bindings, and checks their realization relative to the candidate's reference record. For the numerical channel, $W_h$ contains selected cells, $\mathcal C_h\setminus W_h$ contains captured cells to preserve, and properties outside $\mathcal C_h$ are not covered by that contract. Goal-image comparison occurs only in the ordinary visual verifier. Delivery, shortcuts, and the six-execution cap are specified in Algorithm~\ref{alg:executionrefiner}. Repair construction makes no additional language-model calls.}
\label{fig:loop}
\end{figure}

\section{Method}
\label{sec:method}

\vetr{} refines complete scene programs through a \emph{propose--execute--observe--verify} loop. A private execution refiner constructs alternatives from bounded runtime observations and checks their realization before ordinary visual verification. Runtime acceptance and visual goal assessment serve different purposes: the former checks a candidate's stated effects and preservation conditions; the latter guides subsequent editing toward the goal.

\subsection{Executable World Representation}
\label{subsec:world}

A task supplies a base scene $B$, an editable initial program $p_0$, a start image $r_0$, and a goal image $y^\star$. Following code-based scene representations~\citep{wang2026codeasworld}, the generator submits a complete Python program $q_t$, conditioned on the task and its retained interaction history. Complete means executable relative to the supplied objects and assets in $B$, not a standalone scene constructor. Each submission and repair executes afresh from $B$; the initial program is not run as an extra preparation step before each submission.

The \emph{raw execution record} $z_t=(q_t,r_t,S_t,\tau_t)$ associates the submitted program with its render, captured state, and observed writes. The captured state is $S_t=(\theta_t,\Pi_t,G_t)$, where $\theta_t$ stores supported numerical properties, $\Pi_t$ stores object-to-resource references, and $G_t$ stores captured scene structure. The bounded write trace is
\begin{equation}
  \tau_t=\bigl[(o_i,k_i,v_i,\ell_i)\bigr]_{i=1}^{m_t},
  \label{eq:method-trace}
\end{equation}
where an event records the owner resolved at execution time, a supported property path, its observed value, and its source location. The observer covers supported assignments in the submitted source, including local helpers; imported-library writes and unresolved aliases do not provide owner-qualified events. Unobserved effects remain unknown. Execution failures retain available observations and diagnostics, with unavailable record fields left undefined.

We distinguish $z_t$ from the \emph{delivered record} $\bar z_t$, which may contain a repaired program, the raw result, or the preceding delivered record $z^{-}$. Generated program/render pairs retain their common provenance. Initial observations $(S_0,\tau_0)$ are obtained by executing $p_0$ once without an additional render. We denote the initial reference by $z_0=(p_0,r_0,S_0,\tau_0)$, marking $r_0$ as the supplied visual anchor, not a render from that observation pass. The refiner takes $z_0$ so that both initial writes and their final endpoints are available (Appendix~\ref{app:executionrefiner}).

\subsection{Execution Refiner}
\label{subsec:refiner}

The refiner examines every submitted execution, including successful proposals whose images already improve. Planning actions do not invoke it. It uses $z_t$, the initial observations, and $z^{-}$ to construct and validate alternative complete programs without additional language-model calls.

\paragraph{Candidate effects from execution evidence.}
For a supported non-light cell $(o,k)$ and source execution $s\in\{t,0\}$, define its \emph{displaced-value indices} in $\tau_s$ as the observed writes whose values differ from that source's final endpoint. The recovery endpoint is the latest such value:
\begin{equation}
  \begin{aligned}
    \mathcal I_s(o,k)&=\{i:o_i=o,\ k_i=k,\ v_i\not\simeq\theta_s(o,k)\},\\
    \widehat v_s(o,k)&=v_{\max\mathcal I_s(o,k)}\quad\text{when }\mathcal I_s(o,k)\ne\varnothing.
  \end{aligned}
  \label{eq:method-recovery}
\end{equation}
Here $\simeq$ denotes the runtime checker's numerical agreement predicate. Current and initial traces supply separate recovery hypotheses, both applied to the current complete program. The light-recovery channel similarly reconstructs observed owner-local energy/color requests displaced by later shared writes, resetting earlier requests after explicit data rebinding. These values are alternative endpoints, not inferred user intent or proof that a later assignment was erroneous.

In the ordered controller, if no recovery candidate has been accepted, the light channel considers eligible full and half effective changes between $S^{-}$, the previous delivered state, and $S_t$. For owners in the submitted edit's supported scope, the \emph{projected light endpoint} is
\begin{equation}
  u_{\alpha,o}=\phi(S^{-}(o))+\alpha\bigl[\phi(S_t(o))-\phi(S^{-}(o))\bigr],
  \qquad \alpha\in\{1,\tfrac12\},
  \label{eq:method-projection}
\end{equation}
where $\phi(E,c)=Ec$ is the compiler's effective-value map for light energy $E$ and color $c$. Projection modifies the previous delivered program, retaining its other captured state. It therefore differs from recovery, which modifies the raw proposal. The map supplies algebraic parameter targets, not a guarantee about the rendered image.

\paragraph{Compatible values and resource ownership.}
Each candidate $h$ specifies a reference record $z_h^{\mathrm{ref}}$, required endpoints, and permitted changes. Recovery uses $z_h^{\mathrm{ref}}=z_t$; projection uses $z_h^{\mathrm{ref}}=z^{-}$. The compiler appends a patch to that reference program, producing $q'_h$. It retains reference values for other captured cells and makes permitted local resource copies when shared owners require incompatible endpoints. Compatible sharing is retained. Non-light compilation restores supported values directly; light compilation uses a bounded reconstruction of $\phi$ and compatible owner tuples. Because $\phi$ is not one-to-one, its formula alone does not specify the reconstruction rule (Appendix~\ref{app:executionrefiner}).

\paragraph{Runtime realization and preservation.}
The compiled program executes afresh from $B$. For the numerical-property channel, let $\mathcal C_h$ be the reference's captured supported cells, $W_h\subseteq\mathcal C_h$ the selected write support, and $d_h$ its required numerical endpoints. Fresh readback must satisfy
\begin{equation}
  \theta'_h(c)\simeq
  \begin{cases}
    d_h(c), & c\in W_h,\\
    \theta_h^{\mathrm{ref}}(c), & c\in\mathcal C_h\setminus W_h.
  \end{cases}
  \label{eq:method-frame}
\end{equation}
Additional checks cover captured node/mesh structure, protected object frames, and render context. Light candidates are checked against their captured owner tuples, effective endpoints, and permitted reference changes. The captured cells $\mathcal C_h\setminus W_h$ are known values to preserve; properties outside $\mathcal C_h$ are not covered by this numerical contract. Channel-specific checks constrain the refiner's changes relative to its reference, not every change made by the generator, and are not relaxed after failure.

\paragraph{Validation and delivery.}
The fixed order is current/initial non-light recovery, current/initial light recovery, then full-/half-effect projection. Unsupported alternatives or those lacking required evidence cannot pass the contract. Algorithm~\ref{alg:executionrefiner} makes the six-additional-execution cap explicit and returns the first accepted alternative. It also retains endpoint-based skip rules: raw reuse as full-effect projection requires that candidate's complete runtime contract relative to $z^{-}$. If the endpoint matches but those checks fail, the reported branch skips the compiled full-effect program rather than executing it. Such a skip is not proof of an equivalent rerender; Appendix~\ref{app:executionrefiner} explains its implications. If no alternative is accepted, retain a successful raw result, otherwise reuse the exact preceding record. Raw fallback is not certification under a repair contract.

The ordinary visual verifier then evaluates the delivered result against $y^\star$. Passing a runtime contract neither requires nor guarantees lower image error. The recursion is within-task program revision: subsequent proposals use the delivered result and retained feedback, without a monotonic-improvement guarantee or a claim that each program encodes the entire history.

\subsection{Role-Asymmetric Evidence Organization}
\label{subsec:evidence}

The refiner's internal evidence and the agents' visible contexts are distinct. The refiner consumes observed writes, captured endpoints, and resource bindings. The generator retains its actions and delivered tool outcomes in a bounded multimodal history, with 24 recent messages by default and the task/planning inputs pinned separately. The visual verifier receives the current delivered comparison and returns a goal-directed critique; its default context is reset rather than sharing the generator's accumulated transcript. This organization does not require either agent to interpret the complete private trace. The three-message and shared-memory variants change these context policies (Appendix~\ref{app:components}).

Algorithm~\ref{alg:wide} places execution refinement inside the host loop and distinguishes the generator-interaction cap from additional repair executions. At width $N>1$, each executed candidate is refined before the visual tournament selects a delivered outcome. Final evaluation uses the last successful selected render from a completed run, not the best historical image; failures do not replace its associated program. The supplied start image alone does not constitute a successful generated answer (Appendix~\ref{app:answer}).

\section{Experiments}
\label{sec:eval}

\subsection{Experiment Setup}
\label{sec:protocol}
We evaluate the complete \vetr{} system on two established test suites, BlenderGym~\citep{gu2025blendergym} and BlenderBench~\citep{yin2026viga}, which evaluate program-based 3D graphics editing across diverse visual modification tasks. Throughout the main benchmark results, \vetr{} denotes the full system with execution refinement enabled by default; the otherwise matched variant with refinement disabled is reported only in the ablation study.

\textbf{Tasks and systems.} BlenderGym contains 280 editing tasks: 85 blend-shape, 50 placement, 55 geometry, 45 lighting, and 45 material tasks \citep{gu2025blendergym}. Each task supplies an initial program, start image, and goal image. BlenderBench contains 27 released tasks across three difficulty levels, with nine tasks per level. We compare One-Shot (direct generation from the initial program and visual target without iterative feedback), BlenderAlchemy, VIGA, and \vetr{}. GPT-5.6-luna uses three separate 280-task cohorts: systems, context ablations, and scene origin/guidance. All GPT-5.6-luna and Qwen3.8-27B runs use Blender 3.6.0, and every paired task shares a runtime. VIGA/\vetr{} allow ten generator interactions. Each ablation changes only its named setting; system comparisons retain each system's instructions and execution conventions. Table~\ref{tab:cohortmap} maps these cohorts to the full results and configurations.

\textbf{Measurements.} We report negative CLIP similarity $\nclip=1-\langle\hat f(r),\hat f(y^\star)\rangle$, using unit-normalized CLIP ViT-B/32 embeddings, and photometric loss $\PL$, the RGB mean squared error in $[0,1]$ \citep{radford2021clip}. Lower is better; tables scale both by 100. The same stored answer renders also provide PSNR, SSIM, AlexNet LPIPS, Canny edge F1, and CLIP similarity \citep{wang2004ssim,zhang2018lpips}. For PSNR, SSIM, edge F1, and CLIP similarity, higher is better, whereas for LPIPS, lower is better. PSNR is computed per image before averaging; CLIP and $\nclip$ are algebraically related. Edge F1 uses thresholds 100/200 on 8-bit grayscale and a two-pixel Chebyshev matching tolerance. The evaluated iterative wrapper renders Camera1; the evaluator averages matched views when more than one is available. Appendix~\ref{app:extmetrics} provides all definitions. Paired comparisons use the tasks valid for both systems. For each, we report a 95\% task-level bootstrap confidence interval (5{,}000 resamples of tasks with replacement, seed 0) for the relative reduction in mean error, and an exact two-sided sign test on per-task wins with ties excluded; a difference is \emph{statistically significant} when the interval excludes zero and $p<.05$ (Appendix~\ref{app:paired}).

\subsection{BlenderGym}
\label{sec:main}
We first evaluate the complete \vetr{} system against one-shot generation and iterative baselines across diverse editing families. These system comparisons characterize visual quality under the reported configurations; the component ablations separately test execution refinement and evidence organization within the host loop.

Table~\ref{tab:main} reports the complete-system results. \vetr{} achieves the strongest overall performance on both GPT-5.6-luna and Qwen3.8-27B, and consistently outperforms One-Shot, VIGA, and BlenderAlchemy across the aggregate metrics and nearly all family-level comparisons. On GPT, \vetr{} attains the best $\PL$ and $\nclip$ in four of the five editing families, with material editing as the only exception; on Qwen, it achieves the best result in every reported family--metric cell. These results are particularly notable in light of the original VIGA study~\citep{yin2026viga}, which reported substantial gains over a GPT-4o One-Shot baseline. In our experiments with newer and stronger backbones, however, the gap between iterative visual refinement and One-Shot becomes considerably smaller and can even reverse. A plausible explanation is that improvements in foundation-model capabilities have substantially strengthened one-shot program generation, reducing the headroom available to refinement strategies that rely primarily on rendered visual feedback. In contrast, \vetr{} continues to improve substantially over these stronger One-Shot outputs. Rather than depending on an initially weak program that can be corrected through repeated visual inspection, \vetr{} exploits execution evidence to identify which symbolic edits were actually realized, overwritten, or disrupted, and recursively incorporates this information into subsequent program states. These gains are statistically significant against One-Shot on both backbones: \vetr{} lowers $\nclip$/$\PL$ by 52.7\%/58.8\% on GPT and by 28.1\%/49.6\% on Qwen, and on Qwen it also improves over VIGA by 40.0\%/56.3\%. In each case the bootstrap interval excludes zero and the sign test gives $p\le.0043$ (Appendix~\ref{app:paired}, Table~\ref{tab:paired}).

Table~\ref{tab:ext} complements the primary error metrics with pixel, structural, perceptual, and semantic similarity measures, reporting \vetr{} at $N=4$. On GPT-5.6-luna, its improvements over One-Shot satisfy the paired significance criterion for all five metrics on 214 jointly valid tasks (sign tests $p\le2.2\times10^{-12}$). Against VIGA, edge F1 and LPIPS improve by 5.6\% and 18.5\%, respectively, on 279 jointly valid tasks; both pass the criterion ($p=.0030$ and $.0227$). Thus the gains extend beyond photometric error to edge agreement and learned perceptual similarity. CLIP provides a similarity-scale view of the primary $\nclip$ score, rather than an independent metric (Appendix~\ref{app:extmetrics}, Table~\ref{tab:extpaired}).

On Qwen3.8-27B, \vetr{} has the best reported means for PSNR, SSIM, LPIPS, and CLIP, and ties One-Shot on edge F1 at the displayed precision. These means use each system's valid tasks, so we assess significance on matched subsets (Table~\ref{tab:extpaired}). At $N=4$, all five metrics improve significantly over VIGA on 264 paired tasks; against One-Shot, SSIM, LPIPS, and CLIP meet the criterion on 250 tasks, whereas PSNR and edge F1 do not. No extended metric establishes an advantage of $N=4$ over $N=1$, separating the evidence for refinement from that for increased candidate breadth.

\begin{table}[!htbp]
  \centering
  \small
\caption{\textbf{Visual editing performance on BlenderGym across editing families.}
$\PL$ and $\nclip$ are scaled by 100 ($\downarrow$).
Parentheses indicate negative improvement.
}

  \label{tab:main}

  \setlength{\tabcolsep}{2.8pt}
  \renewcommand{\arraystretch}{1.12}

  \resizebox{\linewidth}{!}{%
  \begin{tabular}{@{}ll*{12}{c}r@{}}
    \toprule
    \multirow{2}{*}{Model}
    & \multirow{2}{*}{Setting}
    & \multicolumn{2}{c}{All}
    & \multicolumn{2}{c}{Shape}
    & \multicolumn{2}{c}{Placement}
    & \multicolumn{2}{c}{Geometry}
    & \multicolumn{2}{c}{Lighting}
    & \multicolumn{2}{c}{Material}
    & \multirow{2}{*}{\shortstack{Impr.\\(\%)$\uparrow$}} \\
    \cmidrule(lr){3-4}
    \cmidrule(lr){5-6}
    \cmidrule(lr){7-8}
    \cmidrule(lr){9-10}
    \cmidrule(lr){11-12}
    \cmidrule(lr){13-14}
    & &
    $\PL\downarrow$ & $\nclip\downarrow$ &
    $\PL\downarrow$ & $\nclip\downarrow$ &
    $\PL\downarrow$ & $\nclip\downarrow$ &
    $\PL\downarrow$ & $\nclip\downarrow$ &
    $\PL\downarrow$ & $\nclip\downarrow$ &
    $\PL\downarrow$ & $\nclip\downarrow$ & \\
    \midrule

    \multirow{4}{*}{GPT-5.6-luna}
    & One-Shot
    & 1.389 & 2.583
    & 0.460 & 1.629
    & 0.955 & 2.966
    & 4.350 & 5.754
    & 1.906 & 1.555
    & 2.109 & \best{5.237}
    & -- \\

    & BlenderAlchemy
    & 5.675 & 9.032
    & 1.176 & 3.567
    & 0.895 & 4.993
    & 10.506 & 14.553
    & 9.969 & 11.421
    & 11.429 & 17.588
    & (236.65) \\

    & VIGA
    & 1.354 & 4.017
    & 0.265 & 0.864
    & 0.551 & 1.608
    & 4.242 & 10.977
    & 0.916 & 1.161
    & \best{1.276} & 7.155
    & 16.92 \\

    & \vetr{}
    & \best{1.034} & \best{2.397}
    & \best{0.246} & \best{0.775}
    & \best{0.527} & \best{1.270}
    & \best{2.914} & \best{4.660}
    & \best{0.495} & \best{0.681}
    & 1.326 & 5.661
    & \best{41.22} \\
    \midrule

    \multirow{4}{*}{Qwen3.8-27B}
    & One-Shot
    & 2.160 & 3.520
    & 0.522 & 1.849
    & 0.923 & 2.820
    & 3.956 & 6.201
    & 1.748 & 1.697
    & 5.978 & 7.412
    & -- \\

    & BlenderAlchemy
    & 2.370 & 4.000
    & 0.654 & 2.132
    & 0.815 & 2.380
    & 6.462 & 7.450
    & 2.292 & 2.366
    & 2.429 & 6.794
    & (9.94) \\

    & VIGA
    & 2.950 & 4.750
    & 0.804 & 1.938
    & 0.760 & 2.390
    & 8.055 & 11.528
    & 2.755 & 2.605
    & 4.585 & 8.192
    & (31.35) \\

    & \vetr{}
    & \best{1.330} & \best{2.910}
    & \best{0.452} & \best{1.375}
    & \best{0.686} & \best{2.115}
    & \best{3.557} & \best{5.666}
    & \best{0.747} & \best{1.080}
    & \best{1.630} & \best{5.225}
    & \best{30.44} \\
    \bottomrule
  \end{tabular}%
  }

\centering\small
\caption{\textbf{Complementary image-similarity metrics.}
Higher is better for PSNR, SSIM, Edge F1, and CLIP;
lower is better for LPIPS.}
\label{tab:ext}
\setlength{\tabcolsep}{4pt}
\renewcommand{\arraystretch}{1.08}

\begin{minipage}[t]{0.49\linewidth}
\centering
\textbf{GPT-5.6-luna}\\[2pt]
\scalebox{0.78}{%
\begin{tabular}{l ccccc}
\toprule
System & PSNR$\uparrow$ & SSIM$\uparrow$
& Edge F1$\uparrow$ & LPIPS$\downarrow$ & CLIP$\uparrow$ \\
\midrule
One-Shot
& 25.17 & 0.884 & 0.773 & 0.127 & 0.974 \\
BlenderAlchemy
& 19.00 & 0.787 & 0.614 & 0.266 & 0.928 \\
VIGA
& 28.18 & 0.911 & 0.812 & 0.087 & 0.978 \\
\vetr{}
& \best{30.85} & \best{0.924} & \best{0.840}
& \best{0.069} & \best{0.988} \\
\bottomrule
\end{tabular}}
\end{minipage}\hfill
\begin{minipage}[t]{0.49\linewidth}
\centering
\textbf{Qwen3.8-27B}\\[2pt]
\scalebox{0.78}{%
\begin{tabular}{l ccccc}
\toprule
System & PSNR$\uparrow$ & SSIM$\uparrow$
& Edge F1$\uparrow$ & LPIPS$\downarrow$ & CLIP$\uparrow$ \\
\midrule
One-Shot
& 23.84 & 0.838 & \best{0.729} & 0.171 & 0.965 \\
BlenderAlchemy
& 22.84 & 0.817 & 0.692 & 0.192 & 0.960 \\
VIGA
& 22.05 & 0.795 & 0.683 & 0.207 & 0.952 \\
\vetr{}
& \best{24.48} & \best{0.841} & \best{0.729}
& \best{0.151} & \best{0.971} \\
\bottomrule
\end{tabular}}
\end{minipage}
\end{table}

\subsection{BlenderBench}
\label{sec:blenderbench-results}
\label{sec:native-refiner-closed-loop}
BlenderBench is a more demanding testbed than BlenderGym, with camera adjustment, multi-step editing, and compositional tasks that introduce stronger dependencies between successive scene modifications. These settings create more opportunities for intermediate errors to propagate, making them particularly suitable for evaluating whether \vetr{} can preserve reliable refinement under increasingly entangled execution trajectories.

Across the 27 released BlenderBench tasks, \vetr{} achieves the best result in every level--metric cell on both GPT-5.6-luna and Qwen3.8-27B, outperforming One-Shot and VIGA throughout the benchmark. Relative to VIGA on the same tasks, \vetr{} reduces $\nclip$/$\PL$ by 55.0\%/67.7\% on GPT and 27.6\%/48.6\% on Qwen, showing that its advantage remains substantial even as task complexity increases.

The strongest separation appears on GPT Level~3, the most compositional setting, where \vetr{} lowers $\nclip$/$\PL$ by 72.4\%/75.2\% relative to VIGA. This is precisely the regime in which execution-grounded refinement is most useful: when multiple edits interact, a failure may arise not only from an incorrect visual judgment but also from symbolic writes that are overwritten, only partially realized, or otherwise altered during execution. By retaining the execution trace and converting witnessed state changes into bounded program updates, \vetr{} can correct these failures before they cascade into later steps. In contrast, render-only refinement has less direct evidence about which intermediate edits actually survived execution.

Every iterative row completes all 27 tasks; GPT One-Shot renders 21/27, and its unrendered tasks are excluded from its reported means.

\begin{table}[!htbp]
\centering\small
\caption{\textbf{Visual editing on native BlenderBench by level.} Improvement is the mean relative reduction from the same-backbone One-Shot row over the six level--metric cells, as defined in Eq.~\eqref{eq:nativeimprovement}. }
\label{tab:native-refiner}
\scalebox{0.78}{%
\begin{minipage}{\dimexpr\textwidth*100/78\relax}
\begin{tabular*}{\linewidth}{@{\extracolsep{\fill}}ll*{6}{w{r}{46pt}}r@{}}
\toprule
Model & Setting & \multicolumn{2}{c}{Level 1} & \multicolumn{2}{c}{Level 2} & \multicolumn{2}{c}{Level 3} & Impr.\\
\cmidrule(lr){3-4}\cmidrule(lr){5-6}\cmidrule(lr){7-8}
 & & \multicolumn{1}{c}{$\PL\downarrow$} & \multicolumn{1}{c}{$\nclip\downarrow$} & \multicolumn{1}{c}{$\PL\downarrow$} & \multicolumn{1}{c}{$\nclip\downarrow$} & \multicolumn{1}{c}{$\PL\downarrow$} & \multicolumn{1}{c}{$\nclip\downarrow$} & \multicolumn{1}{c}{$\uparrow$}\\
\midrule
\multirow{3}{*}{GPT-5.6-luna}
 & One-Shot & 10.9403 & 14.8074 & 2.7780 & 4.4044 & 14.3415 & 26.3542 & ---\\
 & VIGA & 11.2304 & 14.7821 & 5.3481 & 5.0353 & 21.7647 & 22.8812 & $(24.65\%)$\\
 & \vetr{} & \textbf{6.3523} & \textbf{10.1947} & \textbf{0.6519} & \textbf{2.7021} & \textbf{5.3922} & \textbf{6.3201} & $\mathbf{+54.45\%}$\\
\midrule
\multirow{3}{*}{Qwen3.8-27B}
 & One-Shot & 12.9096 & 19.3977 & 3.9448 & 5.1091 & 18.5005 & 36.1053 & ---\\
 & VIGA & 9.4124 & 14.2022 & 4.5780 & 4.8665 & 13.6108 & 15.1571 & $+21.17\%$\\
 & \vetr{} & \textbf{7.7325} & \textbf{11.6530}& \textbf{0.2585} & \textbf{3.9672}& \textbf{6.1927}& \textbf{9.1418}& $\mathbf{+56.17\%}$\\
\bottomrule
\end{tabular*}
\end{minipage}}
\end{table}

The improvement column averages the six level--metric cells against the same-backbone One-Shot row. Let $e_{b,\ell,m}$ denote the error for baseline $b$, level $\ell$, and metric $m$, and let $e_{s,\ell,m}$ denote the corresponding system error. We report
\begin{equation}
\operatorname{Improvement}(s)=\frac{100}{6}
\sum_{\ell=1}^{3}\sum_{m\in\{\nclip,\PL\}}
\left(1-\frac{e_{s,\ell,m}}{e_{b,\ell,m}}\right)\%,
\label{eq:nativeimprovement}
\end{equation}
which weights the three levels and two metrics equally. Each method's level means use only the tasks it rendered.

\begin{figure}[t]
  \centering
  \includegraphics[width=\linewidth]{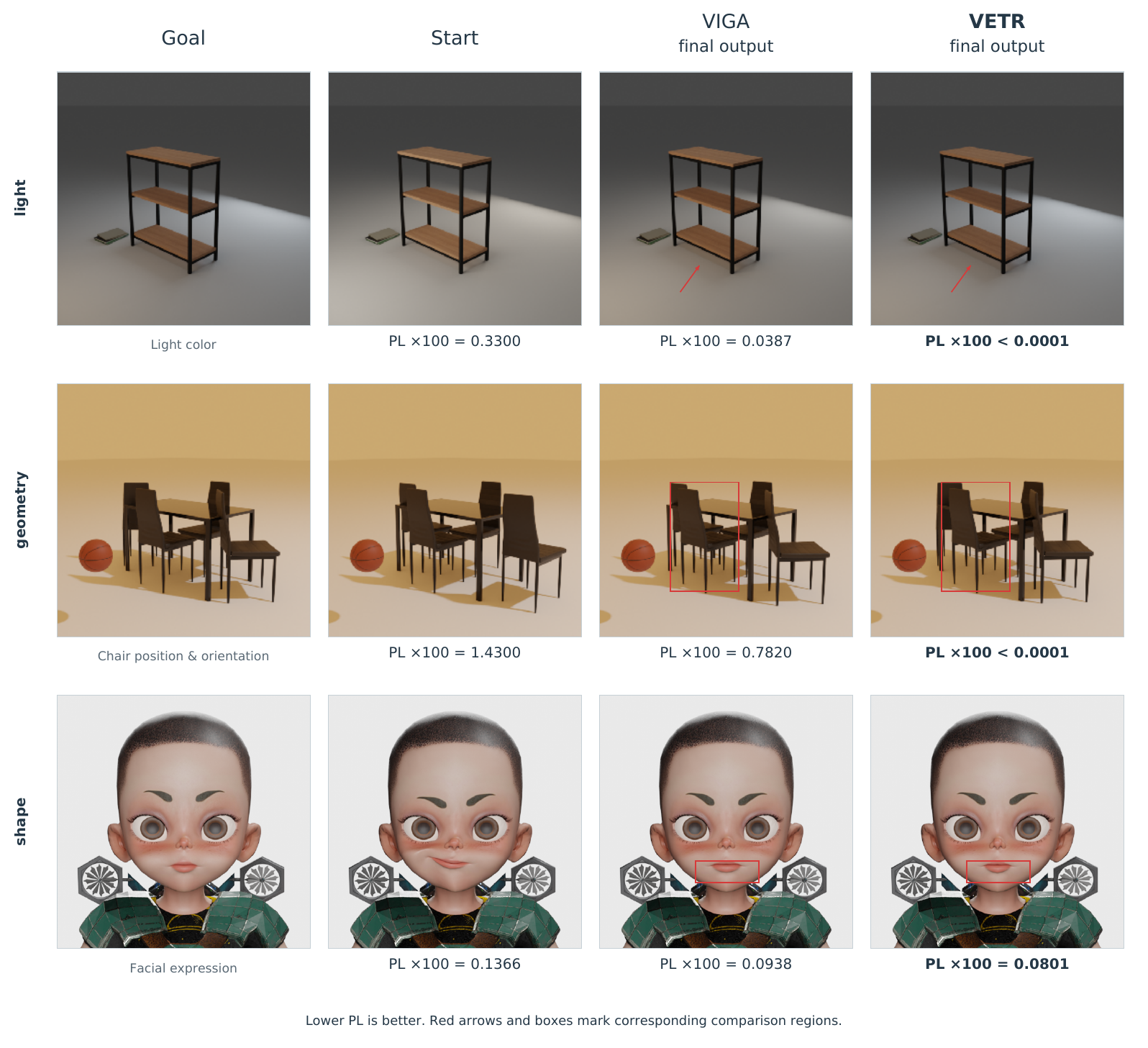}
  \caption{\textbf{Qualitative comparison of visual editing.} Columns display the goal, start, VIGA final output, and VETR final output across three editing categories: light (color), geometry (chair position \& orientation), and shape (facial expression). Red arrows and boxes highlight local regions of comparison. Reported values denote PL $\times 100$ (lower is better). VETR achieves near-zero PL in the displayed light and geometry tasks and lower error in the displayed shape task. These output comparisons illustrate visual fidelity, not an independent verification of the refiner's runtime mechanism.
 }
\label{fig:qualitative-comparison}
\end{figure}

\paragraph{Qualitative comparison.}
Figure~\ref{fig:qualitative-comparison} compares three selected final outputs, not intermediate repair interventions. Values here follow the figure's $\PL\times100$ scale. On the displayed light task, \vetr{} has error $<0.0001$ versus VIGA's $0.0387$; on geometry, $<0.0001$ versus $0.7820$; on shape, $0.0801$ versus $0.0938$. The marked regions illustrate differences in illumination, placement, and facial shape. These examples support the stated final-image comparisons, but do not identify which repair produced a difference or independently verify preservation of scene properties. The refiner's contribution is evaluated by the controlled on/off comparison in Section~\ref{sec:refinement-ablation}.

\subsection{Ablation Study}
\label{sec:ablation}
The ablation study examines three design choices: (i) adding the execution refiner within a fixed generator-interaction cap; (ii) using separate generator and verifier contexts rather than shared memory; and (iii) retaining a longer generator history. These interventions test the refiner and its surrounding evidence interface without assuming that both agents consume the private runtime trace.

\paragraph{Execution refinement.}
\label{sec:refinement-ablation}
We first isolate the execution-refiner stage, which converts observed writes and state transitions into runtime-validated updates. Disabling only this stage holds prompts, tools, base-scene execution policy, backbone, and the ten-generator-interaction cap fixed. The enabled stage can still perform additional repair executions.

Disabling refinement degrades performance in 11 of the 12 level--metric cells across the two backbones (Table~\ref{tab:refinement-ablation}), often by substantial margins. On GPT, removing refinement increases both $\PL$ and $\nclip$ errors at every difficulty level, with the largest degradation on Level~3: $\PL$ rises from 5.3922 to 11.8475 and $\nclip$ from 6.3201 to 15.2108. The same pattern holds on Qwen, where disabling refinement substantially increases errors on Levels~1 and~3 and raises Level~2 $\PL$ from 0.2585 to 2.9924. Paired statistics over the 27 GPT tasks confirm the effect: refinement lowers mean $\PL$ by 41.8\%, which is statistically significant (bootstrap interval $[6.7,49.0]$ excludes zero; sign test 24/3 wins, $p=4.9\times10^{-5}$), and lowers mean $\nclip$ by 32.2\% with a bootstrap interval of $[14.4,55.5]$ (Appendix~\ref{app:refinerpaired}, Table~\ref{tab:refinerpaired}).

These results support adding execution refinement within the same generator-interaction cap. Constructing a repair makes no additional language-model call, but validation can add up to six execution attempts per submitted program. The comparison therefore holds the generator budget fixed, not total computation.

\begin{table}[!htbp]
\centering
\caption{\textbf{Ablation.} Left: execution refinement on the 27 released BlenderBench tasks, using GPT-5.6-luna and Qwen3.8-27B. Right: context ablations on 280 BlenderGym tasks with GPT-5.6-luna, from a separate run in which all rows are paired on the same tasks. Errors are scaled by 100 ($\downarrow$); paired statistics are in Appendices~\ref{app:refinerpaired} and~\ref{app:components}.}

\label{tab:refinement-ablation}
\label{tab:roleviews}
\vspace{2pt}
\small
\mbox{%
\scalebox{0.78}{%
\setlength{\tabcolsep}{2pt}%
\begin{tabular}{@{}ll*{6}{w{r}{36pt}}@{}}
\toprule
Model & Setting & \multicolumn{2}{c}{Level 1} & \multicolumn{2}{c}{Level 2} & \multicolumn{2}{c}{Level 3}\\
\cmidrule(lr){3-4}\cmidrule(lr){5-6}\cmidrule(lr){7-8}
 & & \multicolumn{1}{c}{\scriptsize$\PL\downarrow$} & \multicolumn{1}{c}{\scriptsize$\nclip\downarrow$} & \multicolumn{1}{c}{\scriptsize$\PL\downarrow$} & \multicolumn{1}{c}{\scriptsize$\nclip\downarrow$} & \multicolumn{1}{c}{\scriptsize$\PL\downarrow$} & \multicolumn{1}{c}{\scriptsize$\nclip\downarrow$}\\
\midrule
\multirow{2}{*}{GPT-5.6-luna}
 & \vetr{} (default) & \textbf{6.3523} & \textbf{10.1947} & \textbf{0.6519} & \textbf{2.7021} & \textbf{5.3922}  & \textbf{6.3201}\\
 & w/o refinement & 8.3768 & 10.4058 & 1.0725 & 2.7410 & 11.8475 & 15.2108 \\
\midrule
\multirow{2}{*}{Qwen3.8-27B}
 & \vetr{} (default) & \textbf{7.7325} & \textbf{11.6530}& \textbf{0.2585} & 3.9672& \textbf{6.1927}& \textbf{9.1418}\\
 & w/o refinement & 11.3198 & 13.8822 & 2.9924 & \textbf{3.3747} & 12.3879 & 18.5392 \\
\bottomrule
\end{tabular}}%
\hspace{4pt}%
\scalebox{0.78}{%
\setlength{\tabcolsep}{0pt}%
\renewcommand{\arraystretch}{1.36}%
\begin{tabular}{@{}l@{\hspace{5pt}}w{r}{32pt}@{\hspace{5pt}}w{r}{32pt}@{}}
\toprule
Setting & \multicolumn{1}{w{c}{32pt}}{\scriptsize$\nclip\downarrow$} & \multicolumn{1}{w{c}{32pt}}{\scriptsize$\PL\downarrow$} \\
\midrule
\vetr{} (default)& \best{2.6030}& \best{0.9963} \\
w/o role asymmetry & 2.9347 & 1.2327 \\
w/o visual verifier & 2.7742 & 1.2776 \\
w/o full history & 3.8647 & 1.7838 \\
\bottomrule
\end{tabular}}}
\end{table}

\paragraph{Component-wise ablation.}
We ablate three design choices in one 280-task GPT-5.6-luna cohort, pairing each variant with the default on the same tasks (Table~\ref{tab:roleviews}, right; statistics in Appendix~\ref{app:components}): a memory shared by generator and verifier (\textbf{w/o role asymmetry}), no visual verifier, and a generator limited to its three most recent messages instead of 24 (\textbf{w/o full history}).
Limiting history degrades most, raising $\nclip$/$\PL$ by 48.5\%/79.0\%; both increases are statistically significant, with bootstrap intervals excluding zero and sign tests $p<10^{-5}$, which supports retaining the longer generator history. Removing role asymmetry raises them by 12.7\%/23.7\%, also statistically significant on both metrics (bootstrap intervals exclude zero; sign tests $p\le.003$), supporting separate role contexts. Full statistics are in Appendix~\ref{app:components} (Table~\ref{tab:component}). Removing the verifier raises them by 6.6\%/28.2\%, and the default wins on more tasks (sign tests $p\le.016$; Appendix~\ref{app:components}).

\section{Discussion and Conclusion}
\label{sec:discussion}
\vetr{} treats executions as evidence for revising an executable scene. Authoritative records identify the result being refined; observed writes and state transitions expose how proposed changes become realized effects. The refiner turns selected observations into executable updates by jointly choosing supported values and compatible bindings, then validates the resulting program against runtime contracts before visual verification. On native BlenderBench, adding this execution stage lowers aggregate visual error on both evaluated model backbones. Recorded interventions show how the same evidence-to-program pipeline operates inside ordinary visual reasoning, while the host-loop studies characterize the surrounding execution and context interfaces.

\paragraph{Scope and transfer.}
The BlenderGym history and execution-origin contrasts cover broad target-view editing; the native BlenderBench comparison evaluates the complete closed-loop refiner on released tasks. Runtime validation establishes realization and preservation only for the captured properties. Goal relevance is assessed by the ordinary visual verifier, allowing subsequent proposals to build on the delivered result. For another domain, such as SVG or CAD, the same principle requires observable program effects, supported update operators, and an explicit preservation contract. The current refiner operates on a fixed set of write types derived from the Blender execution model, so extending it to new domains requires defining additional write grammars and preservation boundaries. Future work could explore learned preservation models that infer coupling constraints directly from execution traces.

\bibliographystyle{iclr2027_conference}
\bibliography{vetr}

\subsection*{AI use statement}
Generative AI assisted code inspection, literature organization, manuscript and figure preparation, and statistical scripting. Experimental measurements come from the recorded runs and stored renders. The authors are responsible for the final claims, references, analyses, and artifacts.
 
\subsection*{Reproducibility statement}
The supplement provides execution and context settings, full paired tables, metric definitions, and per-task provenance. The supplied scripts and recorded programs, renders, and outcomes support regeneration of the reported analyses. The native BlenderBench results retain per-level measurements for both model backbones; the GPT records additionally retain per-task scores, task identities, source hashes, and runtime counters for all 27 tasks of every iterative arm and the 21 rendered One-Shot tasks.
\clearpage
\appendix
\raggedbottom

\section{Related Work}

\paragraph{Executable representations and program revision.}
Programs provide compositional control over scene structure: LayoutGPT, 3D-GPT, Holodeck, and SceneCraft generate scenes via language models \citep{feng2023layoutgpt,sun2023threedgpt,yang2024holodeck,hu2024scenecraft}; Scene Language unifies programs with semantic and visual representations \citep{zhang2025scenelang}; Code-as-World treats executable worlds as revisable hypotheses \citep{wang2026codeasworld}. 
NLI4VolVis combines language-guided editing with multi-agent control over editable 3D Gaussian representations~\citep{ai2026nli4volvis}, while end-to-end agent harnesses coordinate planning, implementation, and validation of visual analysis applications~\citep{miao2026aiviscoscientists}.
For iterative editing, BlenderAlchemy, VIGA, and VULCAN combine rendered feedback, visual memory, and geometric constraints \citep{gu2024blenderalchemy,yin2026viga,kuang2026vulcan}; EZBlender adds local validation \citep{wang2026ezblender}; SEIG imposes stage budgets and rollback \citep{he2026thinkingblender}. 
These systems establish feedback loops but lack mechanisms to extract alternatives from execution traces. \vetr{} fills this gap by compiling observed writes and bindings into bounded updates validated before the next proposal.

\paragraph{Agent memory, execution evidence, and preservation.}
Feedback-driven refinement is studied by Self-Refine, Reflexion, CRITIC \citep{madaan2023selfrefine,shinn2023reflexion,gou2024critic}, Self-Debugging, and CodeAct \citep{chen2024selfdebug,wang2024codeact}; VDebugger leverages stepwise traces \citep{wu2024vdebugger}; visual unit tests guide revision \citep{panagopoulou2025viunit}; and agent-memory research examines experience reuse \citep{hu2026memoryagentbench,xu2025amem,zhang2026ace}. 
HiLSVA combines feedback-driven adaptation with stepwise provenance~\citep{ai2026hilsva}, while SciVisAgentBench evaluates agent performance through task outcomes and multimodal assessment~\citep{ai2026scivisagentbench}.
Preservation is a complementary requirement: Prompt-to-Prompt, Plug-and-Play, and MasaCtrl maintain structure or appearance \citep{hertz2023prompt,tumanyan2023plug,cao2023masactrl}; PIE-Bench, EditBench, MagicBrush, and ICE-Bench evaluate preservation alongside editing \citep{ju2024pnp,wang2023editbench,zhang2023magicbrush,pan2025icebench}. 

\paragraph{Execution-guided updates and local reasoning.}
SemFix uses symbolic execution for repair synthesis \citep{nguyen2013semfix}; bidirectional evaluation propagates outputs backward through functional programs \citep{mayer2018bidirectional}. 
Unlike these methods targeting internal values, \vetr{} derives candidates from witnessed imperative executions against an image objective, requiring no additional model inference. 
Separation logic and RustBelt motivate local state reasoning \citep{reynolds2002separation,jung2018rustbelt}; we instantiate this as finite runtime checks over captured properties. 
Graphics systems like Mitsuba 2 and OpenUSD provide the foundation of reusable, interacting scene resources \citep{nimierdavid2019mitsuba,openusdinstancing}, within which \vetr{} performs recursive within-task program revision using execution-grounded evidence.

\section{Execution-refiner implementation details}
\label{app:executionrefiner}
\subsection{Compiling trace evidence into realized edits}
\label{app:executionrefiner-detail}
The execution refiner converts bounded observations into constrained executable updates between raw execution and visual verification (Figure~\ref{fig:loop}). It constructs candidate effects from earlier writes displaced during execution or effective changes between runtime states, then compiles them into parameter updates and permitted resource-ownership changes. The generator still proposes complete programs with its original prompt and tool schema. The refiner examines every submitted execution, including improving raw renders; planning actions do not invoke it. No additional language model identifies an error or constructs a repair.

\paragraph{Observe the write, its owner, and its fate.}
Alongside $\operatorname{Execute}(B,q_t)$, the observer records
\begin{equation}
\tau_t=\bigl[(o_i,k_i,v_i,\ell_i)\bigr]_{i=1}^{m_t},
\qquad S_t=(\theta_t,\Pi_t,G_t).
\label{eq:appendix-observations}
\end{equation}
Here $o_i$ is the owner bound at the executed statement, $k_i$ a supported numerical property path, $v_i$ its observed value, and $\ell_i$ its source location. Final numerical values, object-to-resource references, and captured structure reside in $\theta_t$, $\Pi_t$, and $G_t$, respectively. Resolving owners during execution distinguishes calls to the same submitted helper that write different objects. The observer covers supported assignments in submitted source, including local helpers; imported-library writes and unresolved aliases do not provide owner-qualified candidate evidence. A supported standalone assignment is read back at the next eligible Python trace event, observing application-level coercion without evaluating its right-hand side twice. Unsupported bindings and incomplete observations remain unknown.

\paragraph{Initial reference and required evidence.}
The public initial program $p_0$ is executed once for $(S_0,\tau_0)$ without another render. We write $z_0=(p_0,r_0,S_0,\tau_0)$ for this initial reference, explicitly marking $r_0$ as the supplied visual anchor rather than the output of this observation pass. Initial recovery needs both $\tau_0$ and the initial endpoint map $\theta_0$ in $S_0$; Algorithm~\ref{alg:executionrefiner} therefore takes $z_0$, matching its caller. Subsequent delivered program/render pairs retain the provenance of their generated render. Reusing the initial anchor after failure is not a successful generated answer.

Recovery requires a source write, its source endpoint, and a resolvable target in the reference program. Projection requires the in-scope light observations in both $S_t$ and $S^{-}$. A missing required observation cannot establish a candidate's endpoint or preservation condition; unavailable evidence is not assigned a default value. The exact supported-path, owner-matching, and missing-data policies remain implementation details identified in Section~\ref{app:implementation-boundary}.

\paragraph{Lost values become competing hypotheses.}
For source execution $s\in\{t,0\}$ and a supported non-light cell $(o,k)$, the recovery rule selects
\begin{equation}
\widehat v_s(o,k)=v_{i^\star},\qquad
 i^\star=\max\{i\text{ in }\tau_s:o_i=o,\ k_i=k,\ v_i\not\simeq\theta_s(o,k)\},
\label{eq:appendix-recovery}
\end{equation}
when this set is nonempty. The current and initial traces supply separate source hypotheses. Both propose restoring eligible cells in the current complete program; initial-source recovery does not rewind the whole program to $p_0$. The per-cell rule supplies candidate values, not by itself a complete definition of how eligible cells are grouped into a write support.

The light channel reconstructs observed owner-local energy/color requests displaced by later shared writes, resetting earlier requests after explicit data rebinding. This reset requires ordered binding evidence: the final reference map $\Pi_s$ alone does not describe when a binding changed. Last-write-wins defines the executed result but does not identify the program that best matches a new visual goal. Displaced values remain hypotheses, not inferred intent or proof that a later assignment was erroneous.

\paragraph{Compile values and compatible ownership.}
For a recovery hypothesis, let $W_h$ be the supported cells to change and $d_h(o,k)$ their required endpoints; other captured cells retain their raw endpoints. Shared owners with incompatible requirements require permitted local copying before writes. Compatible requirements do not force a split. A shared geometry-node group, for example, cannot realize two different required radii through one shared value; lamps with compatible requests may retain shared data.

The light compiler uses the effective-value map $\phi(E,c)=Ec$ to enumerate bounded scalar-write supports and group owners by compatible required tuples. Non-light compilation restores supported values directly and detaches conflicting material, mesh, camera, or node-group references where supported. The map $\phi$ does not uniquely determine the light reconstruction: $(E,c)$ and $(2E,c/2)$ have the same effect whenever both are admissible. Consequently, the bounds, selected tuple components, ordering of feasible reconstructions, and allowed copying/rebinding must come from the compiler rather than being inferred from $\phi$.

\paragraph{Project observed effects into executable updates.}
In the ordered controller of Algorithm~\ref{alg:executionrefiner}, projection is considered if no recovery candidate has been accepted, subject to the projection channel's evidence and scope requirements. For each in-scope owner $o$, Eq.~\eqref{eq:method-projection} forms $u_{\alpha,o}$ using $S^{-}$ and $S_t$, with $\alpha=1$ before $\alpha=\tfrac12$. These endpoints come from executions, not hidden target parameters. Projection compiles into the previous delivered program while retaining its other captured state; recovery instead uses the raw program. Thus, candidate construction, compilation, and the decision to execute a compiled program are distinct steps. The endpoint-based shortcuts below can skip a constructed candidate.

\paragraph{Preserve a bounded frame around the repair.}
Each compiled program runs afresh from $B$, not a scene mutated by a previous trial. For numerical-property \emph{recovery}, the reference is the raw record and the check is
\begin{equation}
\theta'_h(o,k)\simeq
\begin{cases}
 d_h(o,k), &(o,k)\in W_h,\\
 \theta_t(o,k), &(o,k)\in\mathcal C_h\setminus W_h.
\end{cases}
\label{eq:appendix-recovery-frame}
\end{equation}
This is Eq.~\eqref{eq:method-frame} specialized to recovery, not a different reference rule for projection. The captured complement $\mathcal C_h\setminus W_h$ is known and checked; properties outside $\mathcal C_h$ are not covered by this numerical contract. Further checks retain captured node/mesh structure, protected object frames, and render context. Light candidates are checked against their owner tuples, effective endpoints, and allowed reference changes. Every check uses the candidate's designated reference: raw for recovery, previous delivered for projection. This finite operational frame follows the local-reasoning distinction between changed and retained state~\citep{reynolds2002separation,jung2018rustbelt}; it is not a guarantee about uncaptured properties or visual quality.

\paragraph{A single acceptance contract, including raw reuse.}
Write $\mathsf{Valid}_h(z')$ for the conjunction of successful execution, a readable render, realization of all required endpoints, preservation relative to $z_h^{\mathrm{ref}}$, and applicable scope/binding checks. Required unknown values cannot make this predicate true. It is the same acceptance obligation for a fresh trial and for reusing the raw record as a projection candidate. In particular, raw reuse under full-effect projection must satisfy preservation relative to $z^{-}$, not merely agree with its own numerical endpoints. This notation specifies the contract; it does not supply undocumented numerical tolerances or an implementation of the checks.

\paragraph{Endpoint shortcuts and the skipped full-effect candidate.}
The supplied controller skips a light-recovery trial when its required endpoint is already realized by the raw record. For full-effect projection, it returns the raw record only if the endpoint-match test and $\mathsf{Valid}_h(z)$ both pass. If the endpoint matches but the contract fails, it \emph{continues to the next candidate without executing the compiled full-effect program}. Algorithm~\ref{alg:executionrefiner} retains this branch; it is not silently changed to fall through to execution.

The distinction is substantive. Under the effective-endpoint interpretation,
\[
 u_{1,o}=\phi(S^{-}(o))+[\phi(S_t(o))-\phi(S^{-}(o))]=\phi(S_t(o)).
\]
Hence, whenever these values are available, the raw light effect matches by construction. If the shortcut tests only this effect, the full-effect stage acts as a raw-reuse check or skip, rather than a fresh full-effect repair. A check of a more restrictive required owner tuple is a different predicate and must be specified explicitly. We do not infer such a tuple check from the effective-value equation.

As a schematic example, let a previous record have light energy $1$ and a protected object position $x=0$, with fixed light color. A raw proposal with energy $2$ and $x=5$ realizes the full-effect light target, but violates a projection that must retain $x=0$. A compiled program with energy $2$ and $x=0$ is not equivalent to the raw program. Skipping it is therefore a restriction of the search policy, not proof that rerendering would be redundant. The endpoint-match predicates need implementation confirmation (Section~\ref{app:implementation-boundary}).

\paragraph{Execution cap, delivery, and visual feedback.}
The order is current/initial non-light recovery, current/initial light recovery, then full-/half-effect projection. Supported alternatives are constructed on demand. The counter $b$ in Algorithm~\ref{alg:executionrefiner} makes the reported limit explicit: at most six \emph{additional execution attempts} per submitted program, including failed attempts. Enumerating scalar supports during compilation is not itself an execution. The counter does not assume that each channel necessarily supplies exactly one alternative.

A successful fresh trial passing $\mathsf{Valid}_h$ is returned immediately. When no candidate is accepted, a successful raw execution is retained; execution/render failure returns the exact preceding delivered record. Raw fallback is not a claim that the raw program passed a repair contract. Diagnostics for a failed submission are kept distinct from a cached render's provenance. The ordinary visual verifier subsequently compares the delivered program/image with the goal. No image-error decrease is required by runtime acceptance, and no additional language-model call constructs the repair.

\begin{algorithm}[tbp]
\caption{Execution-evidence refinement: reported skip policy and explicit execution cap}
\label{alg:executionrefiner}
\begin{algorithmic}[1]
\Require base $B$, raw record $z$, initial reference $z_0$, previous delivered record $z^{-}$
\State $b\gets0$ \Comment{additional executions for this submission}
\State Inspect available writes, endpoints, binding evidence, and supported edit scope
\For{each supported alternative $h$ in the fixed recovery/projection order}
  \If{a required observation or supported compilation is unavailable}
    \State \textbf{continue}
  \EndIf
  \State Set $z_h^{\mathrm{ref}}\gets z$ for recovery, or $z^{-}$ for projection
  \State Compile $q'_h$ and its endpoint, preservation, scope, and binding requirements
  \If{$h$ is light recovery and its endpoint-match predicate holds on $z$}
    \State \textbf{continue} \Comment{reported endpoint-based skip}
  \EndIf
  \If{$h$ is full-effect projection and its endpoint-match predicate holds on $z$}
    \If{$\mathsf{Valid}_h(z)$} \Comment{full projection contract relative to $z^{-}$}
      \State \Return $z$
    \EndIf
    \State \textbf{continue} \Comment{skip $q'_h$; do not fall through to execution}
  \EndIf
  \If{$b=6$}
    \State \textbf{break}
  \EndIf
  \State $b\gets b+1$; $z'\gets\textsc{ExecuteAndTrace}(B,q'_h)$
  \If{$\mathsf{Valid}_h(z')$}
    \State \Return $z'$
  \EndIf
\EndFor
\If{the raw program executes successfully and produces a readable render}
  \State \Return $z$
\EndIf
\State \Return $z^{-}$ \Comment{reuse the exact preceding record}
\end{algorithmic}
\end{algorithm}

\paragraph{Specification boundary.}
\label{app:implementation-boundary}
The equations and listing above clarify the recorded procedure but do not fully specify its implementation-dependent selectors. Exact reproduction still requires the supported-property registry, source-to-reference owner matching, selection/grouping of $W_h$, and the rule selecting projection scope. The light inverse additionally requires admissible bounds, required tuple components, feasible-solution ordering, allowed copies/rebindings, and infeasibility handling. Ordered rebinding observations cannot be recovered from the final $\Pi_s$ alone; their recording and alignment with numerical writes must be specified. Finally, the numerical agreement predicate, missing-readback behavior, shortcut endpoint tests, and recovery-to-projection eligibility must be tied to the evaluated code. No undocumented values, selectors, or tie-breaking rules are assumed here. In particular, the ordered presentation of projection after unaccepted recovery must not be substituted for a different eligibility gate in the actual controller.

\paragraph{Making an edit observable.}
Let $R_B(q)$ denote the rendered result of a complete program. Distinct programs can satisfy $q_1\ne q_2$ yet $R_B(q_1)\simeq R_B(q_2)$ when execution erases their differing intermediate values. Their traces may still supply different alternatives. Conversely, multiple parameterizations of a light effect can differ in surrounding state. Execution evidence supplies bounded hypotheses whose realization can be tested; the visual verifier assesses their goal relevance. An improving raw image does not establish that every observed request survived, motivating examination of every submitted execution without asserting that each repair improves the image.

\makeatletter\setlength{\@fptop}{0pt}\makeatother
\renewcommand{\topfraction}{0.9}\renewcommand{\bottomfraction}{0.8}\renewcommand{\textfraction}{0.07}\renewcommand{\floatpagefraction}{0.8}
\setcounter{topnumber}{3}\setcounter{bottomnumber}{2}\setcounter{totalnumber}{4}

\section{Execution-interface diagnostics and cost}
\label{app:interfacediagnostics}
\subsection{Diagnostic study: consistent realization of proposals}
\label{sec:scene}
To diagnose execution consistency under the complete-program interface, this study retains full generator history in both arms and changes the scene loaded for execution. Rebuild loads the task's initial scene each time; persistent scene loads and saves a job-local scene across rounds. Both use GPT, $N=1$, the same verifier, a complete-program interface, and ten generator interactions. This separate run set completes 278/280 tasks in each arm, with 276 jointly valid answers.
 
Rebuilding reduces mean $\nclip$ by 33.9\%, with a 95\% interval of $[19.8,45.8]$ and 154 wins, 120 losses, and two ties. The effect is concentrated in geometry, with a 53.4\% reduction; Appendix~\ref{app:scene} reports every family.

A deterministic replay isolates one mechanism. Executing the same saved \texttt{geometry20} program nine times on a persistent scene accumulates nine modifiers and reduces mesh volume to $1/256$ of its first-execution value. Rebuild keeps one modifier and constant volume. Explicitly replacing previous modifiers on the persistent scene also prevents accumulation (Appendix~\ref{app:replay}). Thus, the requirement is consistency between a program and its execution origin; a compatible incremental or replacement interface can also retain scene state.
 
\subsection{Refinement cost and candidate breadth}
\label{sec:cost}
With an interaction cap $T$ and candidate width $N$, at most $TN$ raw candidate programs enter the execution stage. Each submission permits at most $R=6$ additional repair executions, so the raw-program/repair stage has a worst-case bound of $(1+R)TN$ execution attempts, plus the separate initial-program observation. This is an upper bound rather than a reconstruction of recorded cost counters. Planning actions, unsupported candidates, failures, and no-op shortcuts can reduce execution or render counts. A tournament over $k$ available delivered renders uses $k-1$ pairwise comparisons. The refiner itself makes no additional language-model calls; the generator, ordinary verifier, and tournament have their own inference costs. The generator's visible history is bounded by its message window, while archived trajectories grow with recorded interactions.
 
\begin{table}[tbp]
\centering\small
\caption{\textbf{Recorded cost in the GPT Blender 3.6 cohort.} Means use tasks with saved cost records ($n$); completion is reported separately. Quality for both widths is reported in Table~\ref{tab:breadth}.}
\label{tab:costmain}
\setlength{\tabcolsep}{3.5pt}
\resizebox{\ifdim\width>\linewidth\linewidth\else\width\fi}{!}{%
\begin{tabular}{@{}lrrrrr@{}}
\toprule
Variant & $n$ & Interactions & Model calls & Blender renders & Agent time (s) \\
\midrule
\vetr{} $N=1$ & 277 & 8.74 & 18.58 & 7.16 & 204 \\
\vetr{} $N=4$ & 280 & 8.76 & 66.17 & 28.56 & 733 \\
\bottomrule
\end{tabular}}
\end{table}
 
Table~\ref{tab:costmain} makes the cost of wider search explicit: $N=4$ uses about $3.6\times$ the model calls and $4.0\times$ the renders of $N=1$. The core contrasts use $N=1$.

\section{Complete refinement procedure}
\label{app:algorithm}
Algorithm~\ref{alg:wide} expands the single-submission procedure to both candidate widths. At $N=1$, action voting and the visual tournament reduce to the single available response. At $N>1$, only candidates of the selected action type are executed. Each render-producing candidate passes through Algorithm~\ref{alg:executionrefiner} before visual candidate selection. All candidates in an interaction use the same previous delivered record; only the selected outcome advances that record. The tournament compares delivered candidates, whereas runtime acceptance inside the refiner uses no visual-improvement gate. Planning and inspection actions consume generator interactions but do not invoke the refiner.

\begin{algorithm}[tbp]
\caption{Complete refinement procedure with optional candidate selection}
\label{alg:wide}
\begin{algorithmic}[1]
\Require task inputs $x=(B,p_0,r_0,y^\star)$, interaction cap $T$, width $N$, history window $L_G$
\State Initialize generator history $H$ with its prompt and task inputs
\State Obtain $z_0$ by observing $p_0$ without rerendering; set $z^{-}\gets z_0$
\For{$j=1,\ldots,T$}
  \State Sample $N$ actions from the generator's window $W_G(H;L_G)$
  \State Parse actions and choose the most frequent valid action type
  \If{no valid action}
    \State Append parsing feedback; log candidates; \textbf{continue}
  \EndIf
  \If{the chosen action produces renders}
    \For{each candidate $a$ of the selected type, with complete program $q_a$}
      \State $z_a\gets\textsc{ExecuteAndTrace}(B,q_a)$
      \State $\bar z_a\gets\textsc{Refine}(B,z_a,z_0,z^{-})$ \Comment{Algorithm~\ref{alg:executionrefiner}}
    \EndFor
    \State Select one delivered outcome by its render, otherwise use diagnostics
    \State Set $z^{-}$ to the selected delivered record, when available
  \Else
    \State Execute one instance of the chosen non-render action
  \EndIf
  \If{no outcome is returned}
    \State Append failure feedback; log candidates; \textbf{continue}
  \EndIf
  \State Obtain visual-verifier feedback when requested by the selected tool
  \State Append selected action and delivered outcome to $H$; log all candidates
  \If{the chosen action is end} \State \textbf{break} \EndIf
\EndFor
\State \textbf{Evaluation:} score the last successful selected generated render
\Statex \hspace{\algorithmicindent}under Appendix~\ref{app:answer}; cached images retain their original provenance
\end{algorithmic}
\end{algorithm}

The returned record and failure diagnostics have distinct provenance. If an attempted script fails and the refiner returns a cached record, the cached render remains paired with its own program, not with the failed submission. The supplied initial image is a reference, not a successful generated render. Runs with no successful generated render, and timed-out or incomplete runs, remain invalid under the existing answer rule.

\section{Complete system results}
\label{app:systems}
Table~\ref{tab:main} reports family-level means and completion counts in the main text. The complete paired comparisons in Appendix~\ref{app:paired} provide the head-to-head analyses. The descriptive improvement column in Table~\ref{tab:main} is
\begin{equation}
\operatorname{Impr}(m)=\frac{100}{10}\sum_{k=1}^{10}\frac{x_k^{\mathrm{One\text{-}Shot}}-x_k^m}{x_k^{\mathrm{One\text{-}Shot}}},
\label{eq:impr}
\end{equation}
where $k$ indexes the five families and two benchmark metrics. GPT uses the single Blender 3.6 system cohort, including BlenderAlchemy with four candidates per interaction. Qwen retains its separately documented configuration (Appendix~\ref{app:qwen}).

\paragraph{GPT completion and paired quality.}
One-Shot completes 214 tasks; failures concentrate in geometry (23/55 valid) and material (13/45). The $N=1$ comparison therefore uses the 212 tasks completed by both systems, and $N=4$ uses all 214 One-Shot tasks. Table~\ref{tab:oneshotpaired} reports the matched One-Shot reference for each intersection. Relative reductions are 40.3\%/56.3\% ($\nclip$/$\PL$) for $N=1$ and 52.7\%/58.8\% for $N=4$. Both variants improve each extended metric on their corresponding pairs.
 
\paragraph{Comparison with iterative systems.}
Against VIGA, $N=1$ reduces $\nclip$ by 36.3\% (CI $[24.3,46.4]$, 161 wins and 114 losses, $p=.0054$). At $N=4$, the mean reductions are 40.6\% in $\nclip$ and 24.6\% in $\PL$. Both widths improve both primary metrics relative to BlenderAlchemy.
 
\paragraph{Qwen: gains over VIGA extend to a second backbone.}
On Qwen3.8-27B, $N{=}1$ reduces $\nclip$/$\PL$ by 31.9\%/50.8\% on 261 paired tasks, and $N{=}4$ by 40.0\%/56.3\% on 264; all four comparisons meet both criteria.
Against One-Shot, $N{=}4$ establishes 28.1\%/49.6\% reductions (250 tasks), and $N{=}1$ yields 22.1\%/43.6\% (248 tasks).
Both widths outperform BlenderAlchemy on both metrics.
Completion is 251/280 for One-Shot, 265 for VIGA, 279 for BlenderAlchemy, 276 for $N{=}1$, and 279 for $N{=}4$.
 
\paragraph{Additional image metrics.}
Table~\ref{tab:ext} summarizes image similarity, while Table~\ref{tab:extpaired} reports matched effects, confidence intervals, and win counts. On GPT-5.6-luna, both \vetr{} variants improve all five metrics significantly over One-Shot; against VIGA, the $N=1$ CLIP comparison and the $N=4$ edge F1 and LPIPS comparisons meet the paired criterion. On Qwen3.8-27B, $N=1$ establishes gains in PSNR, LPIPS, and CLIP over VIGA, and in LPIPS over One-Shot. At $N=4$, all five comparisons with VIGA and the SSIM, LPIPS, and CLIP comparisons with One-Shot meet the criterion. Neither backbone establishes an extended-metric advantage of $N=4$ over $N=1$
 
\section{Editing strategies}
\label{sec:strategy}
\begin{figure}[tbp]
\centering\footnotesize
\begin{minipage}[t]{0.49\linewidth}\raggedright
\textbf{(a)} Start program (\texttt{geometry28}, 86 lines; excerpt)\\[1pt]
\begin{lstlisting}[style=prog]
from infinigen.core.nodes.node_wrangler import Nodes
from infinigen.core import surface
 
def shader_material_002_002(nw):
    bsdf = nw.new_node(Nodes.PrincipledBSDF,
      input_kwargs={'Base Color': (0, .0086, .8, 1)})
    ...
def geometry_nodes(nw):
    subdiv = nw.new_node(Nodes.SubdivideMesh,
      input_kwargs={'Mesh': ..., 'Level': 7})
    noise = nw.new_node(Nodes.NoiseTexture)
    ramp = nw.new_node(Nodes.ColorRamp,
      input_kwargs={'Fac': noise.outputs["Fac"]})
    ramp.color_ramp.elements[0].position = 0.0
    ramp.color_ramp.elements[1].position = 1.0
    delete = nw.new_node(Nodes.DeleteGeometry,
      input_kwargs={'Geometry': subdiv,
                'Selection': ramp.outputs["Color"]})
apply(bpy.context.active_object)
\end{lstlisting}
\end{minipage}\hfill
\begin{minipage}[t]{0.49\linewidth}\raggedright
\textbf{(b)} GPT-5.6-luna One-Shot: 293-line rewrite, no render\\[1pt]
\begin{lstlisting}[style=prog]
import bpy, math, random
from mathutils import Vector
# Reset the scene geometry ...
<<for obj in list(bpy.data.objects):>>
<<    if obj.type == "MESH":>>
<<        bpy.data.objects.remove(obj, do_unlink=True)>>
mat = bpy.data.materials.new("FragmentBlue")
nodes = mat.node_tree.nodes
<<nodes.clear()>>
bsdf = nodes.new("ShaderNodeBsdfPrincipled")
bsdf.inputs["Base Color"].default_value = ...
<<bsdf.inputs["Specular IOR Level"].default_value = .28>>
# ... 240 lines: fragments built by hand with bpy
scene = bpy.context.scene
<<scene.render.engine = "BLENDER_EEVEE_NEXT">>
\end{lstlisting}
Retains 4\% of the start lines. Blender 3.6 raises \texttt{KeyError: "Specular IOR Level" not found} at the incompatible socket binding; there is no second attempt, so the task is lost.
\end{minipage}

\vspace{3pt}
\begin{minipage}[t]{0.49\linewidth}\raggedright
\textbf{(c)} GPT-5.6-luna inside \vetr{} $N{=}1$: rounds 1--9\\[1pt]
\begin{lstlisting}[style=prog]
# round 1 -- verifier: "volume not perforated";
#   smallest safe test: lower the subdivision level
    subdiv = nw.new_node(Nodes.SubdivideMesh,
!!      input_kwargs={'Mesh': ..., 'Level': 4})!!
# round 2 -- render unchanged: the noise has no
#   spatial input, so wire object position into it
!!    pos = nw.new_node(Nodes.InputPosition)!!
    noise = nw.new_node(Nodes.NoiseTexture,
!!    input_kwargs={'Vector': pos.outputs["Position"]})!!
# rounds 3-9 -- move the two ramp stops, one per round
!!    ramp.color_ramp.elements[0].position = 0.40!!
!!    ramp.color_ramp.elements[1].position = 0.55!!
\end{lstlisting}
Round-1 program retains 99\% of the start lines; 9 of 9 rounds render; answer $\nclip$ 14.1 ($N{=}4$: 8.7; VIGA: 16.8).
\end{minipage}
\caption{\textbf{Editing strategy and execution outcome on the same task.} (a) The supplied procedural program. (b) Under the One-Shot prompt, GPT-5.6-luna rewrites the scene and accesses a socket absent in Blender 3.6, so this attempt yields no render. Orange highlights the rewrite and incompatible identifiers. (c) With \vetr{}'s prompt and iterative feedback, GPT preserves most of the program and makes the edits highlighted in blue. Comments summarize observed changes and feedback; ellipses omit unchanged or lengthy code. Scores are $\nclip\times100$ ($\downarrow$).}
\label{fig:strategy}
\end{figure}
 
The examples below characterize the archived program trajectories under the One-Shot prompt.
On the 100 geometry and material tasks, the One-Shot runner and Blender 3.6 runtime are fixed. We characterize the observed editing behavior by the fraction of initial-program lines retained in the generated program (Appendix~\ref{app:strategy}, Table~\ref{tab:strategy}).
GPT-5.6-luna retains a median $10\%$ of the start lines: 92 of 100 programs keep less than half, 46 begin by deleting the scene's objects or clearing the node tree, 65 contain a binding that exists only in Blender 4.x (\texttt{"Specular IOR Level"} 44, \texttt{"Lacunarity"} 13, \texttt{BLENDER\_EEVEE\_NEXT} 12); 66 fail, 40 of them with such a binding as the first exception.
Figure~\ref{fig:strategy}(b) shows this strategy on one task.
The observed strategy changes with the full configuration: inside \vetr{}'s loop, whose prompt asks for one to three edits over bindings present in the program or supplied by the verifier, GPT-5.6-luna's \emph{first-round} program retains a median $96\%$ of the start lines on the same 100 tasks (89 above $90\%$), 4 first-round programs contain a 4.x binding, 2 fail to render, and the line completes 100/100 (Figure~\ref{fig:strategy}(c)).
 
\section{Complete context ablations}
\label{app:components}
The context ablations form one cohort: Blender 3.6.0, GPT-5.6-luna, $N{=}1$, ten generator interactions, and 280 attempted tasks per arm. The default run and three variants differ only in the named switch, and quality metrics were computed only after sampling had finished. Every arm has a valid answer on all 280 tasks, so each comparison is paired on the full benchmark. Table~\ref{tab:component} reports the paired statistics, computed with the procedure of Appendix~\ref{app:paired}.

\begin{table}[tbp]
\centering\small
\caption{\textbf{Context ablations in one cohort} (GPT-5.6-luna, $N{=}1$, Blender 3.6.0, 280 tasks). Means are scaled by 100. Red.: relative reduction of the default mean error against the ablated mean (positive favors the default); the main text instead expresses each ablated error relative to the default. CI: 95\% task bootstrap (5,000 resamples, seed 0). W/L/T: default wins/losses/ties. $p$: exact two-sided sign test excluding ties. $\dagger$: CI excludes zero and $p<.05$.}
\label{tab:component}
\setlength{\tabcolsep}{3pt}
\resizebox{\ifdim\width>\linewidth\linewidth\else\width\fi}{!}{%
\begin{tabular}{@{}llrrrccr@{}}
\toprule
Variant & Metric & $n$ & Default & Ablated & Red.\ (\%) [CI] & W/L/T & $p$\\
\midrule
w/o full history (3 msgs) & $\nclip$ & 280 & 2.603 & 3.865 & $+32.7$ \ci{22.5}{41.8}$^\dagger$ & 177/99/4 & $3.10\!\times\!10^{-6}$\\
 & $\PL$ & 280 & 0.996 & 1.784 & $+44.2$ \ci{25.7}{56.9}$^\dagger$ & 183/97/0 & $3.08\!\times\!10^{-7}$\\
\midrule
w/o role asymmetry & $\nclip$ & 280 & 2.603 & 2.935 & $+11.3$ \ci{3.7}{23.5}$^\dagger$ & 164/113/3 & $.0026$\\
 & $\PL$ & 280 & 0.996 & 1.233 & $+19.2$ \ci{2.9}{31.7}$^\dagger$ & 169/108/3 & $.00030$\\
\midrule
w/o visual verifier & $\nclip$ & 280 & 2.603 & 2.774 & $+6.2$ \ci{-5.6}{17.0} & 159/118/3 & $.016$\\
 & $\PL$ & 280 & 0.996 & 1.278 & $+22.0$ \ci{-4.2}{40.8} & 161/117/2 & $.0098$\\
\bottomrule
\end{tabular}}
\end{table}

\paragraph{Execution history.}
Limiting the generator to its three most recent messages produces the largest effect: the default has 32.7\%/44.2\% lower $\nclip$/$\PL$ than the ablated arm, equivalently the ablated errors are 48.5\%/79.0\% higher than the default. Both metrics meet the paired criterion.

\paragraph{Role asymmetry.}
Sharing one memory between the generator and verifier raises both errors. The default wins on 164 and 169 of the 280 tasks, and both metrics meet the paired criterion.

\paragraph{Visual verification.}
Removing the verifier raises both mean errors, and the default wins on more tasks (159 and 161; $p=.016$ and $.0098$).

\section{Scene-origin details and additional trajectories}
\label{app:scenetrajectories}
\begin{figure}[tbp]
\centering
\includegraphics[width=\linewidth]{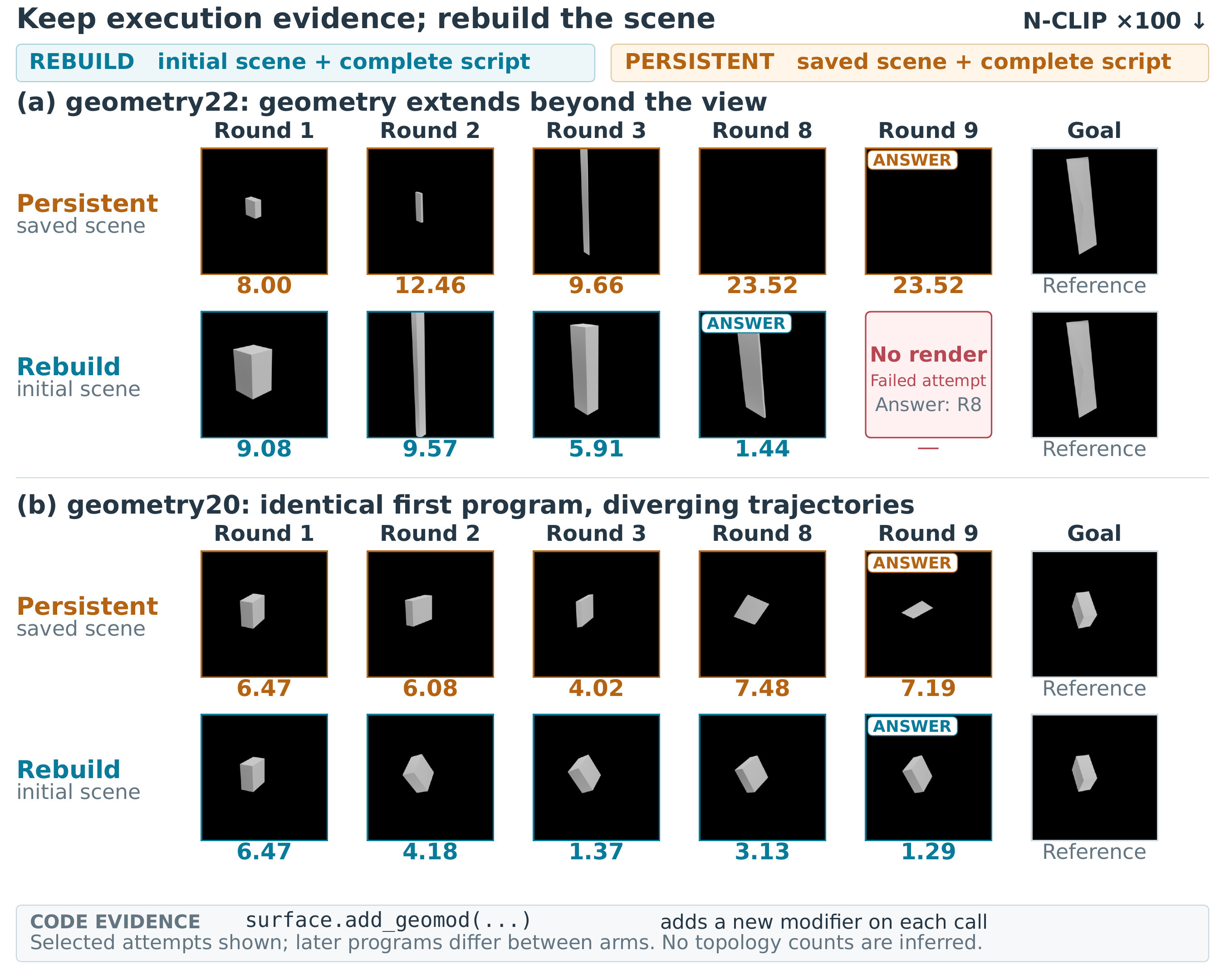}
\caption{\textbf{A complete program can compound edits left in a persistent scene.} Selected execution attempts from the scene-origin ablation; both arms retain the full generator history. (a) On \texttt{geometry22}, the persistent trajectory elongates the object until it leaves the visible frame, while rebuild retains a target-like object. Rebuild's round 9 has no render; its scored answer is round 8 ($\nclip\times100=1.44$), versus persistent's round 9 ($23.52$). (b) On \texttt{geometry20}, the first programs are byte-identical and produce equal scores; later persistent renders become progressively flatter, while rebuild finishes closer to the goal ($1.29$ vs.\ $7.19$). The shared \texttt{surface.add\_geomod} helper creates a new modifier on each call. Appendix~\ref{app:replay} separately executes the identical first program repeatedly and measures modifier counts, mesh volume, and object bounds.}
\label{fig:scene}
\end{figure}

Table~\ref{tab:scene} and Figure~\ref{fig:scene} give the aggregate origin contrast and its recorded trajectories. Both arms retain history; only the scene supplied to complete-program execution changes.
\begin{table}[tbp]
\centering\small
\caption{\textbf{Scene origin matters under complete-program execution.} Both arms retain generator history; only the scene-persistence setting is changed. Means use jointly valid tasks and are scaled by $100$. Red.: reduction of rebuild relative to persistent; CI: 95\% task bootstrap (5,000 resamples, seed 0). W/L/T: rebuild wins/losses/ties. $p$: exact two-sided sign test excluding ties; $\dagger$: CI excludes zero and $p<.05$. Both arms complete 278/280 tasks. Family results are in Appendix~\ref{app:scene}.}
\label{tab:scene}
\setlength{\tabcolsep}{3pt}
\begin{tabular}{@{}l r l r r c c r@{}}
\toprule
Tasks & $n$ & Metric $\downarrow$ & Rebuild & Persistent & Red.\ (\%) [CI] & W/L/T & $p$ \\
\midrule
Overall & 276 & $\nclip$ & 2.371 & 3.585 & $+33.9$ \ci{19.8}{45.8}$^\dagger$ & 154/120/2 & $.0460$ \\
Overall & 276 & $\PL$ & 1.047 & 1.417 & $+26.1$ \ci{-2.4}{43.6} & 147/129/0 & $.3062$ \\
\midrule
Geometry & 52 & $\nclip$ & 4.821 & 10.341 & $+53.4$ \ci{34.7}{67.2}$^\dagger$ & 35/16/1 & $.0110$ \\
Geometry & 52 & $\PL$ & 3.111 & 5.039 & $+38.3$ \ci{-3.4}{61.1} & 38/14/0 & $.0012$ \\
\bottomrule
\end{tabular}
\end{table}
\label{app:scene}
This independent GPT-5.6-luna run set uses all 280 tasks, the full generator context, one candidate per interaction, and a ten-interaction cap.
Both arms request complete scripts and use the same final-answer rule.
The persistent-scene switch copies the initial \texttt{.blend} to a job-local path and uses that path for both loading and saving; the rebuild arm loads the initial task file without saving over it.
Successful scene changes persist only in the first arm.
The execution script saves the scene after execution and rendering.
The two arms are separate sampled trajectories with the same inference settings.
 
\begin{table}[tbp]
\centering\small
\caption{\textbf{Scene-origin ablation by task family.} Same definitions and joint criterion as Table~\ref{tab:scene}. Both means use the paired task set; means are $\times100$. Intervals bootstrap the relative reduction of mean error.}
\label{tab:scenefamily}
\setlength{\tabcolsep}{3pt}
\begin{tabular}{@{}l r l r r c c r@{}}
\toprule
Family & $n$ & Metric $\downarrow$ & Rebuild & Persistent & Red.\ (\%) [CI] & W/L/T & $p$ \\
\midrule
Blend-shape & 84 & $\nclip$ & 0.934 & 0.883 & $(5.8)$ \ci{-49.2}{24.5} & 43/40/1 & $.8264$ \\
Blend-shape & 84 & $\PL$ & 0.301 & 0.313 & $+4.0$ \ci{-26.3}{26.9} & 42/42/0 & $1.0000$ \\
\midrule
Placement & 50 & $\nclip$ & 1.487 & 1.490 & $+0.2$ \ci{-25.7}{21.3} & 26/24/0 & $.8877$ \\
Placement & 50 & $\PL$ & 0.603 & 0.580 & $(3.9)$ \ci{-19.6}{12.0} & 22/28/0 & $.4799$ \\
\midrule
Geometry & 52 & $\nclip$ & 4.821 & 10.341 & $+53.4$ \ci{34.7}{67.2}$^\dagger$ & 35/16/1 & $.0110$ \\
Geometry & 52 & $\PL$ & 3.111 & 5.039 & $+38.3$ \ci{-3.4}{61.1} & 38/14/0 & $.0012$ \\
\midrule
Lighting & 45 & $\nclip$ & 0.773 & 0.765 & $(1.1)$ \ci{-56.1}{36.1} & 25/20/0 & $.5515$ \\
Lighting & 45 & $\PL$ & 0.414 & 0.399 & $(3.8)$ \ci{-64.9}{32.9} & 22/23/0 & $1.0000$ \\
\midrule
Material & 45 & $\nclip$ & 4.800 & 5.966 & $+19.5$ \ci{0.8}{36.0} & 25/20/0 & $.5515$ \\
Material & 45 & $\PL$ & 1.180 & 1.241 & $+4.9$ \ci{-28.5}{24.7} & 23/22/0 & $1.0000$ \\
\bottomrule
\end{tabular}
\end{table}
 
Rebuild has no valid answer on \texttt{blendshape40} and \texttt{geometry53}; persistent has none on \texttt{geometry32} and \texttt{geometry37}.
The intersection therefore contains 84 blend-shape, 50 placement, 52 geometry, 45 lighting and 45 material tasks.
Both arms' stored final renders were scored together with the same evaluator, without agent calls or scene rerendering.
The bootstrap follows the main paired implementation: sorted task IDs, 5,000 samples with replacement using seed 0, and percentile endpoints of the relative mean reduction; exact ties are excluded from the two-sided sign test.
On the persistent-relative scale, the overall and geometry $\nclip$ reductions are 33.9\% and 53.4\%; on the rebuild-relative scale, persistent increases error by 51.2\% and 114.5\%, respectively.
 
\paragraph{Reading the trajectory figure.}
Figure~\ref{fig:scene} displays selected execution attempts 1, 2, 3, 8 and 9; the ten-interaction budget also includes the initial planning action.
For \texttt{geometry22}, rebuild's ninth script fails to produce a render, so its answer is the eighth render.
For \texttt{geometry20}, the first submitted programs have identical SHA-256 hashes.
The helper's modifier allocation is verified in the implementation.
 
\section{Candidate breadth}
\label{sec:breadth}
 
Table~\ref{tab:breadth} compares the two candidate widths in the current GPT cohort. Completion is 278/280 at $N=1$ and 280/280 at $N=4$. On the 278 shared tasks, mean $\nclip$ decreases by 7.7\% (CI $[-4.6,18.4]$, $p=.278$) and $\PL$ by 9.4\% (CI $[-14.7,26.6]$, $p=.436$). The $N=4$ tournament selects a candidate other than candidate zero in 497 of 2,453 interactions (20.3\%). The observed costs are $3.6\times$ the model calls, $4.0\times$ the renders, and $3.6\times$ the agent time of $N=1$. The controlled history result uses the default $N=1$ setting.
\begin{table}[tbp]
\centering\small
\caption{\textbf{Candidate breadth in the GPT Blender 3.6 cohort.} Means use each variant\textquotesingle s valid tasks; the paired row uses their intersection. Costs are in Table~\ref{tab:costmain}. The saved candidate logs contain 2,453 interactions for $N=4$, with 497 nonzero winners; selected indices 0/1/2/3 occur 1,955/222/194/81 times.}
\label{tab:breadth}
\setlength{\tabcolsep}{3.5pt}
\resizebox{\ifdim\width>\linewidth\linewidth\else\width\fi}{!}{%
\begin{tabular}{@{}lrcc@{}}
\toprule
Variant & Valid & $\nclip\downarrow$ & $\PL\downarrow$ \\
\midrule
\vetr{} $N=1$ & 278 & 2.616 & 1.149 \\
\vetr{} $N=4$ & 280 & 2.397 & 1.034 \\
\midrule
Paired red. ($n=278$) & & $+7.7$ \ci{-4.6}{18.4} & $+9.4$ \ci{-14.7}{26.6} \\
\bottomrule
\end{tabular}}
\end{table}
 
\section{Detailed paired comparisons}
\label{app:paired}
Table~\ref{tab:paired} reports current GPT system comparisons and all ten Qwen method pairs. Each comparison pairs valid task identities within its backbone. Table~\ref{tab:oneshotpaired} shows the GPT quality means for each One-Shot intersection.
 
\begin{table}[tbp]
\centering\small
\caption{\textbf{Matched-task GPT image quality.} Each system is paired with One-Shot on their valid-task intersection; the One-Shot reference is recomputed for that exact set. The benchmark errors are scaled by 100. Paired effects and tests are in Tables~\ref{tab:paired} and \ref{tab:extpaired}.}
\label{tab:oneshotpaired}
\setlength{\tabcolsep}{3.5pt}
\resizebox{\ifdim\width>\linewidth\linewidth\else\width\fi}{!}{%
\begin{tabular}{@{}lrrrrrrrr@{}}
\toprule
System (paired subset) & $n$ & $\nclip\downarrow$ & $\PL\downarrow$ & PSNR$\uparrow$ & SSIM$\uparrow$ & Edge F1$\uparrow$ & LPIPS$\downarrow$ & CLIP$\uparrow$ \\
\midrule
One-Shot (N1 pairs) & 212 & 2.607 & 1.402 & 24.67 & 0.883 & 0.771 & 0.129 & 0.974 \\
\vetr{} $N=1$ & 212 & 1.555 & 0.613 & 29.15 & 0.923 & 0.833 & 0.076 & 0.984 \\
\addlinespace[3pt]
One-Shot (N4 pairs) & 214 & 2.583 & 1.389 & 25.17 & 0.884 & 0.773 & 0.127 & 0.974 \\
\vetr{} $N=4$ & 214 & 1.220 & 0.573 & 30.85 & 0.924 & 0.840 & 0.069 & 0.988 \\
\addlinespace[3pt]
One-Shot (VIGA pairs) & 214 & 2.583 & 1.389 & 25.17 & 0.884 & 0.773 & 0.127 & 0.974 \\
VIGA & 214 & 2.198 & 0.877 & 28.18 & 0.911 & 0.812 & 0.087 & 0.978 \\
\addlinespace[3pt]
One-Shot (Alchemy pairs) & 212 & 2.570 & 1.296 & 25.23 & 0.886 & 0.780 & 0.125 & 0.974 \\
BlenderAlchemy & 212 & 7.167 & 4.072 & 19.00 & 0.787 & 0.614 & 0.266 & 0.928 \\
\bottomrule
\end{tabular}}
\end{table}
\begin{table}[tbp]
\centering\small
\caption{\textbf{Detailed paired comparisons by backbone.} N1/N4: \vetr{} $N{=}1/4$; Alchemy: BlenderAlchemy ($2\times2$ for Qwen). $n$: tasks valid for both systems; Red.: relative reduction of the first system\textquotesingle s mean error against the second (positive favors the first); CI: 95\% task-bootstrap interval (5,000 resamples, seed 0); Wins: strictly lower error; $p$: exact two-sided sign test, excluding ties. $\dagger$: CI excludes zero and $p<.05$ (unadjusted). Qwen includes all ten method pairs; BlenderAlchemy uses depth 2, breadth 2, one try, and disabled thinking (Appendix~\ref{app:qwen}).}
\label{tab:paired}
\setlength{\tabcolsep}{3pt}
\resizebox{\ifdim\width>\linewidth\linewidth\else\width\fi}{!}{%
\begin{tabular}{l r rrr rrr}
\toprule
 & & \multicolumn{3}{c}{$\nclip$} & \multicolumn{3}{c}{$\PL$} \\
\cmidrule(lr){3-5}\cmidrule(lr){6-8}
Comparison & $n$ & Red.\,(\%) [95\% CI] & Wins & $p$ & Red.\,(\%) [95\% CI] & Wins & $p$ \\
\midrule
\multicolumn{8}{l}{\textbf{GPT-5.6-luna}} \\
N1 vs.\ VIGA & 277 & $+36.3$ \ci{24.3}{46.4}$^\dagger$ & 161 & $.0054$ & $+16.7$ \ci{-9.1}{36.6} & 149 & $.2294$ \\
N4 vs.\ VIGA & 279 & $+40.6$ \ci{30.2}{49.3} & 149 & $.1845$ & $+24.6$ \ci{5.6}{40.0} & 152 & $.1506$ \\
N1 vs.\ One-Shot & 212 & $+40.3$ \ci{29.4}{50.5}$^\dagger$ & 159 & $8.41\!\times\!10^{-14}$ & $+56.3$ \ci{44.6}{66.7}$^\dagger$ & 144 & $7.70\!\times\!10^{-8}$ \\
N4 vs.\ One-Shot & 214 & $+52.7$ \ci{44.9}{60.3}$^\dagger$ & 168 & $1.19\!\times\!10^{-18}$ & $+58.8$ \ci{45.9}{68.8}$^\dagger$ & 156 & $2.19\!\times\!10^{-12}$ \\
N4 vs.\ N1 & 278 & $+7.7$ \ci{-4.6}{18.4} & 147 & $.2777$ & $+9.4$ \ci{-14.7}{26.6} & 146 & $.4356$ \\
VIGA vs.\ One-Shot & 214 & $+14.9$ \ci{-6.9}{34.1} & 153 & $4.58\!\times\!10^{-11}$ & $+36.8$ \ci{18.3}{51.5}$^\dagger$ & 147 & $4.69\!\times\!10^{-8}$ \\
N1 vs.\ Alchemy & 260 & $+72.5$ \ci{67.1}{77.3}$^\dagger$ & 234 & $5.08\!\times\!10^{-43}$ & $+83.3$ \ci{77.8}{87.5}$^\dagger$ & 222 & $8.49\!\times\!10^{-33}$ \\
N4 vs.\ Alchemy & 262 & $+75.6$ \ci{71.0}{79.7}$^\dagger$ & 241 & $1.54\!\times\!10^{-48}$ & $+83.8$ \ci{79.5}{87.4}$^\dagger$ & 227 & $1.25\!\times\!10^{-35}$ \\
VIGA vs.\ Alchemy & 261 & $+59.7$ \ci{51.3}{67.6}$^\dagger$ & 228 & $4.92\!\times\!10^{-37}$ & $+76.9$ \ci{69.8}{82.4}$^\dagger$ & 217 & $1.17\!\times\!10^{-28}$ \\
Alchemy vs.\ One-Shot & 212 & $(178.9)$ \ci{-238.3}{-133.0}$^\dagger$ & 39 & $2.45\!\times\!10^{-21}$ & $(214.1)$ \ci{-314.4}{-141.4}$^\dagger$ & 56 & $4.18\!\times\!10^{-12}$ \\
\midrule
\multicolumn{8}{l}{\textbf{Qwen3.8-27B}} \\
N4 vs.\ N1 & 275 & $+11.7$ \ci{-1.8}{23.8} & 148 & $.2277$ & $+11.5$ \ci{-6.7}{28.5} & 141 & $.7176$ \\
N4 vs.\ VIGA & 264 & $+40.0$ \ci{30.2}{49.0}$^\dagger$ & 171 & $1.82\!\times\!10^{-6}$ & $+56.3$ \ci{42.4}{66.9}$^\dagger$ & 169 & $6.16\!\times\!10^{-6}$ \\
N4 vs.\ One-Shot & 250 & $+28.1$ \ci{16.0}{39.0}$^\dagger$ & 148 & $.0043$ & $+49.6$ \ci{31.8}{62.6}$^\dagger$ & 149 & $.0023$ \\
N4 vs.\ Alchemy & 278 & $+29.5$ \ci{18.2}{40.0}$^\dagger$ & 186 & $1.80\!\times\!10^{-8}$ & $+44.5$ \ci{30.6}{56.1}$^\dagger$ & 174 & $2.36\!\times\!10^{-5}$ \\
N1 vs.\ VIGA & 261 & $+31.9$ \ci{20.1}{42.4}$^\dagger$ & 162 & $1.16\!\times\!10^{-4}$ & $+50.8$ \ci{34.3}{63.0}$^\dagger$ & 152 & $.0092$ \\
N1 vs.\ One-Shot & 248 & $+22.1$ \ci{9.4}{32.7} & 139 & $.0653$ & $+43.6$ \ci{21.2}{59.4} & 137 & $.0979$ \\
N1 vs.\ Alchemy & 275 & $+17.5$ \ci{4.6}{28.6}$^\dagger$ & 170 & $1.06\!\times\!10^{-4}$ & $+37.2$ \ci{18.5}{50.7}$^\dagger$ & 163 & $.0020$ \\
Alchemy vs.\ VIGA & 264 & $+19.2$ \ci{7.5}{29.8} & 133 & $.9509$ & $+19.9$ \ci{-8.2}{39.0} & 129 & $.7584$ \\
Alchemy vs.\ One-Shot & 250 & $(0.7)$ \ci{-15.7}{12.3} & 105 & $.0186$ & $+8.0$ \ci{-28.7}{35.6} & 119 & $.6108$ \\
VIGA vs.\ One-Shot & 238 & $(15.0)$ \ci{-35.6}{2.1} & 109 & $.2180$ & $(14.2)$ \ci{-63.2}{22.5} & 111 & $.3309$ \\
\bottomrule
\end{tabular}}
\end{table}
 
\paragraph{Paired statistics for the execution-refinement ablation.}
\label{app:refinerpaired}
Table~\ref{tab:refinerpaired} pairs the default \vetr{} run with the refinement-disabled run on the 27 released BlenderBench tasks for GPT-5.6-luna, using the archived per-task ledger; both arms complete all 27 tasks. Reductions compare mean errors across tasks, and intervals, wins, and sign tests follow the procedure above. Refinement lowers mean $\PL$ by 41.8\% (24/3 wins, $p=4.9\times10^{-5}$), meeting the paired criterion, and lowers mean $\nclip$ by 32.2\%, with a bootstrap interval of $[14.4,55.5]$ that excludes zero.
 
\begin{table}[tbp]
\centering\small
\caption{\textbf{Paired effect of execution refinement on BlenderBench (GPT-5.6-luna, 27 tasks).} Means are scaled by 100. Red.: relative reduction of the default mean error against the refinement-disabled run (positive favors refinement); CI: 95\% task bootstrap (5,000 resamples, seed 0); W/L: per-task wins/losses of the default (no ties); $p$: exact two-sided sign test; $\dagger$: CI excludes zero and $p<.05$.}
\label{tab:refinerpaired}
\setlength{\tabcolsep}{4pt}
\begin{tabular}{@{}lrrcrr@{}}
\toprule
Metric & w/o refinement & \vetr{} & Red.\ (\%) [CI] & W/L & $p$\\
\midrule
$\nclip$ & 9.453 & 6.406 & $+32.2$ \ci{14.4}{55.5} & 18/9 & $.122$\\
$\PL$ & 7.099 & 4.132 & $+41.8$ \ci{6.7}{49.0}$^\dagger$ & 24/3 & $4.9\!\times\!10^{-5}$\\
\bottomrule
\end{tabular}
\end{table}
 
\paragraph{Paired statistics for the BlenderBench system comparison.}
\label{app:benchpaired}
Table~\ref{tab:benchpaired} pairs \vetr{} with VIGA on the 27 released BlenderBench tasks for GPT-5.6-luna, using the same archived per-task ledger. Both systems complete all 27 tasks, so the means equal the averages of the level means in Table~\ref{tab:native-refiner}. \vetr{} lowers mean $\nclip$ by 55.0\% and mean $\PL$ by 67.7\%, and both bootstrap intervals exclude zero.

\begin{table}[tbp]
\centering\small
\caption{\textbf{Paired comparison with VIGA on BlenderBench (GPT-5.6-luna, 27 tasks).} Means are scaled by 100. Red.: relative reduction of the \vetr{} mean error against VIGA; CI: 95\% task bootstrap (5,000 resamples, seed 0); W/L: per-task wins/losses of \vetr{} (no ties); $p$: exact two-sided sign test; $\dagger$: CI excludes zero and $p<.05$.}
\label{tab:benchpaired}
\setlength{\tabcolsep}{4pt}
\begin{tabular}{@{}lrrcrr@{}}
\toprule
Metric & VIGA & \vetr{} & Red.\ (\%) [CI] & W/L & $p$\\
\midrule
$\nclip$ & 14.233 & 6.406 & $+55.0$ \ci{29.3}{69.8} & 19/8 & $.052$\\
$\PL$ & 12.781 & 4.132 & $+67.7$ \ci{37.7}{74.1} & 15/12 & $.701$\\
\bottomrule
\end{tabular}
\end{table}
 
\section{Qwen3.8-27B evaluation: settings and scope}
\label{app:qwen}
\paragraph{Backbone and runtime.}
The second-backbone runs use a locally served Qwen endpoint, denoted Qwen3.8-27B, on all 280 tasks. The run environment uses four H100 80\,GB GPUs, Blender 3.6.0, and BlenderGym's Infinigen fork, Cycles CUDA rendering with 512 samples, and the same CLIP ViT-B/32 and RGB-MSE evaluator. One-Shot, VIGA, and \vetr{} use enabled thinking; VIGA and \vetr{} have a ten-round budget.
 
\paragraph{Budgets and failures.}
VIGA and \vetr{} $N{=}1$ have 7,200\,s timeouts; $N{=}4$ and BlenderAlchemy have 21,600\,s. A timeout remains a failure even if an intermediate render exists. One-Shot has 29 no-render failures; VIGA has three timeouts and 12 no-render failures; $N{=}1$ has four timeouts; $N{=}4$ has one (\texttt{geometry37}). BlenderAlchemy's only failure (\texttt{material5}) produced no candidate render after an upstream planning error.
 
\paragraph{BlenderAlchemy configuration.}
The Qwen line uses the released implementation with family-specific edit styles and refinement strategies (depth 2, breadth 2, one try). Visual imagination is disabled because the target render is supplied, and thinking is disabled to obtain parseable planning responses.
 
\paragraph{Statistical scope and traceability.}
All means use successful renders from normally completed tasks. Paired comparisons intersect valid task IDs, bootstrap tasks 5,000 times (seed 0), and apply an exact two-sided sign test after removing ties. Tests are reported without a multiple-comparison correction, as in the primary evaluation. The export contains 1,400 task--method records in \path{per_task_metrics.csv} and full-precision merged records in \path{raw/qwen38_all__metrics_committed.json}, under \path{supplement/blendergym_tables_export/}. Table~\ref{tab:main}'s Qwen block and Table~\ref{tab:paired}'s Qwen block have their tabulated values independently recomputed and numerically cross-checked by \path{supplement/update_results.py}.
 
\section{Extended image metrics}
\label{app:extmetrics}
\paragraph{Definitions.}
The extended scorer reads the same stored final-answer images used for the primary metrics, matches render/goal filenames, and averages per-image values over available views and then over valid tasks.
Images are loaded as RGB in $[0,1]$ at goal resolution; a size mismatch is resized to that resolution.
\emph{PSNR} is $-10\log_{10}\mathrm{MSE}$ in dB, with 100\,dB used when $\mathrm{MSE}\le10^{-10}$.
It is computed per image before averaging.
\emph{SSIM} uses \texttt{skimage.metrics.structural\_similarity}, RGB channels, \texttt{data\_range=1}, and the default uniform $7\times7$ window.
\emph{LPIPS} uses the AlexNet backbone with the package\textquotesingle s default learned calibration and inputs scaled to $[-1,1]$.
\emph{Edge F1} applies OpenCV Canny to both 8-bit grayscale images with thresholds $(100,200)$.
Each edge map is dilated by a $5\times5$ all-ones kernel: a match is within two pixels in Chebyshev distance.
Precision is the fraction of predicted edges matched to a goal edge, and recall the converse; their harmonic mean is reported.
Both maps empty gives 1; exactly one empty gives 0.
\emph{CLIP} is cosine similarity from ViT-B/32 image embeddings, equal to $1-\nclip$ before the latter's $\times100$ table scaling. All CLIP tables derive it from the same canonical per-task $\nclip$ score, so their favorable orderings and ties agree exactly; separately recomputed cosine scores are retained in the scoring exports.
For a higher-is-better metric, paired relative improvement is $(\bar m_{\mathrm{first}}-\bar m_{\mathrm{second}})/\bar m_{\mathrm{second}}$; for LPIPS its sign is reversed. Reported percentages multiply these ratios by 100.
Task-bootstrap CIs, wins, and sign tests follow Appendix~\ref{app:paired}.
 
\paragraph{One-Shot means.}
One-Shot extended metrics cover all 214 of its valid GPT tasks. Its family composition differs from that of iterative systems because failures concentrate in geometry and material. GPT comparisons use 212 jointly valid tasks for $N=1$ and 214 for $N=4$, with each reference recomputed on the same set (Table~\ref{tab:oneshotpaired}).

\begin{table}[tbp]
\centering\small
\caption{\textbf{Paired comparisons on the extended metrics.}
N1/N4 denote \vetr{} at $N=1/4$.
$n$: paired sample size; $\Delta$: PSNR difference in dB and relative improvement (\%) for the other metrics, in the first system\textquotesingle s favorable direction;
CI: 95\% task bootstrap (5{,}000 resamples, seed 0);
W/L: per-task wins/losses (ties excluded); $p$: exact two-sided sign test (unadjusted);
$\dagger$: recorded CI excludes zero and $p<.05$.}
\label{tab:extpaired}
\setlength{\tabcolsep}{2pt}
\renewcommand{\arraystretch}{1.0}
\footnotesize
\resizebox{\linewidth}{!}{%
\begin{tabular}{@{}lrrrr@{\hspace{12pt}}lrrrr@{}}
\toprule
\multicolumn{5}{l}{\textbf{GPT-5.6-luna}} & \multicolumn{5}{l}{\textbf{Qwen3.8-27B}} \\
\cmidrule(lr){1-5}\cmidrule(lr){6-10}
\makebox[32pt][l]{Metric} & \makebox[16pt][r]{$n$} & \makebox[96pt][r]{$\Delta$ [CI]} & \makebox[34pt][r]{W/L} & \makebox[50pt][r]{$p$} &
\makebox[32pt][l]{Metric} & \makebox[16pt][r]{$n$} & \makebox[96pt][r]{$\Delta$ [CI]} & \makebox[34pt][r]{W/L} & \makebox[50pt][r]{$p$} \\
\midrule
\multicolumn{5}{@{}l}{\textit{N1 vs.\ VIGA}} & \multicolumn{5}{@{}l}{\textit{N1 vs.\ VIGA}} \\
PSNR & 277 & $+1.38$ \ci{0.13}{2.64} & 149/128 & $.2294$ & PSNR & 261 & $+1.51$ \ci{0.35}{2.75}$^\dagger$ & 152/109 & $.0092$ \\
SSIM & 277 & $+2.7$ \ci{1.1}{4.4} & 149/125 & $.1646$ & SSIM & 261 & $+4.1$ \ci{2.0}{6.3} & 146/114 & $.0543$ \\
Edge F1 & 277 & $+5.1$ \ci{1.9}{8.6} & 135/110 & $.1250$ & Edge F1 & 261 & $+4.5$ \ci{2.0}{6.9} & 141/120 & $.2157$ \\
LPIPS & 277 & $+14.1$ \ci{5.7}{22.0} & 150/127 & $.1861$ & LPIPS & 261 & $+24.0$ \ci{16.3}{31.3}$^\dagger$ & 155/106 & $.0029$ \\
CLIP & 277 & $+1.5$ \ci{0.9}{2.3}$^\dagger$ & 161/114 & $.0054$ & CLIP & 261 & $+1.6$ \ci{1.2}{2.0}$^\dagger$ & 162/99 & $.0001$ \\
\addlinespace[2pt]
\multicolumn{5}{@{}l}{\textit{N4 vs.\ VIGA}} & \multicolumn{5}{@{}l}{\textit{N4 vs.\ VIGA}} \\
PSNR & 279 & $+2.24$ \ci{1.01}{3.54} & 152/127 & $.1506$ & PSNR & 264 & $+2.48$ \ci{1.34}{3.72}$^\dagger$ & 169/95 & $6.16\!\times\!10^{-6}$ \\
SSIM & 279 & $+2.6$ \ci{0.9}{4.3} & 149/127 & $.2061$ & SSIM & 264 & $+6.3$ \ci{4.1}{8.8}$^\dagger$ & 161/102 & $.0003$ \\
Edge F1 & 279 & $+5.6$ \ci{1.9}{9.6}$^\dagger$ & 144/97 & $.0030$ & Edge F1 & 264 & $+6.9$ \ci{4.7}{9.2}$^\dagger$ & 164/100 & $9.82\!\times\!10^{-5}$ \\
LPIPS & 279 & $+18.5$ \ci{10.6}{26.3}$^\dagger$ & 159/120 & $.0227$ & LPIPS & 264 & $+28.6$ \ci{21.9}{35.1}$^\dagger$ & 173/91 & $5.06\!\times\!10^{-7}$ \\
CLIP & 279 & $+1.7$ \ci{1.1}{2.4} & 149/126 & $.1845$ & CLIP & 264 & $+2.0$ \ci{1.5}{2.4}$^\dagger$ & 171/93 & $1.82\!\times\!10^{-6}$ \\
\addlinespace[2pt]
\multicolumn{5}{@{}l}{\textit{N1 vs.\ One-Shot}} & \multicolumn{5}{@{}l}{\textit{N1 vs.\ One-Shot}} \\
PSNR & 212 & $+4.48$ \ci{2.87}{6.20}$^\dagger$ & 144/66 & $7.70\!\times\!10^{-8}$ & PSNR & 248 & $+0.17$ \ci{-1.28}{1.56} & 137/110 & $.0979$ \\
SSIM & 212 & $+4.6$ \ci{3.2}{6.3}$^\dagger$ & 150/59 & $2.51\!\times\!10^{-10}$ & SSIM & 248 & $+0.3$ \ci{-1.6}{2.2} & 129/117 & $.4832$ \\
Edge F1 & 212 & $+8.1$ \ci{5.4}{11.2}$^\dagger$ & 147/47 & $3.45\!\times\!10^{-13}$ & Edge F1 & 248 & $+0.9$ \ci{-1.3}{3.1} & 132/116 & $.3409$ \\
LPIPS & 212 & $+40.9$ \ci{32.5}{49.1}$^\dagger$ & 162/50 & $5.52\!\times\!10^{-15}$ & LPIPS & 248 & $+15.8$ \ci{6.8}{24.0}$^\dagger$ & 144/104 & $.0131$ \\
CLIP & 212 & $+1.1$ \ci{0.7}{1.4}$^\dagger$ & 159/52 & $8.41\!\times\!10^{-14}$ & CLIP & 248 & $+0.8$ \ci{0.3}{1.3} & 139/109 & $.0653$ \\
\addlinespace[2pt]
\multicolumn{5}{@{}l}{\textit{N4 vs.\ One-Shot}} & \multicolumn{5}{@{}l}{\textit{N4 vs.\ One-Shot}} \\
PSNR & 214 & $+5.68$ \ci{3.94}{7.59}$^\dagger$ & 156/55 & $2.19\!\times\!10^{-12}$ & PSNR & 250 & $+1.37$ \ci{-0.06}{2.79} & 149/100 & $.0023$ \\
SSIM & 214 & $+4.5$ \ci{3.0}{6.2}$^\dagger$ & 157/52 & $1.92\!\times\!10^{-13}$ & SSIM & 250 & $+2.0$ \ci{0.2}{3.9}$^\dagger$ & 146/102 & $.0062$ \\
Edge F1 & 214 & $+8.7$ \ci{5.5}{12.0}$^\dagger$ & 149/44 & $1.52\!\times\!10^{-14}$ & Edge F1 & 250 & $+2.6$ \ci{0.4}{4.8} & 140/110 & $.0664$ \\
LPIPS & 214 & $+45.7$ \ci{38.5}{52.7}$^\dagger$ & 168/46 & $1.61\!\times\!10^{-17}$ & LPIPS & 250 & $+19.8$ \ci{11.8}{27.5}$^\dagger$ & 150/100 & $.0019$ \\
CLIP & 214 & $+1.4$ \ci{1.1}{1.8}$^\dagger$ & 168/43 & $1.19\!\times\!10^{-18}$ & CLIP & 250 & $+1.0$ \ci{0.6}{1.4}$^\dagger$ & 148/102 & $.0043$ \\
\addlinespace[2pt]
\multicolumn{5}{@{}l}{\textit{N4 vs.\ N1}} & \multicolumn{5}{@{}l}{\textit{N4 vs.\ N1}} \\
PSNR & 278 & $+0.88$ \ci{-0.53}{2.28} & 146/132 & $.4356$ & PSNR & 275 & $+1.17$ \ci{-0.19}{2.60} & 141/134 & $.7176$ \\
SSIM & 278 & $-0.1$ \ci{-1.4}{1.2} & 140/134 & $.7627$ & SSIM & 275 & $+1.9$ \ci{0.2}{3.9} & 146/129 & $.3346$ \\
Edge F1 & 278 & $+0.6$ \ci{-2.6}{3.6} & 125/116 & $.6064$ & Edge F1 & 275 & $+3.1$ \ci{0.8}{5.3} & 148/127 & $.2277$ \\
LPIPS & 278 & $+5.1$ \ci{-3.1}{12.6} & 138/140 & $.9522$ & LPIPS & 275 & $+6.9$ \ci{-0.8}{14.7} & 144/131 & $.4694$ \\
CLIP & 278 & $+0.2$ \ci{-0.1}{0.5} & 147/128 & $.2777$ & CLIP & 275 & $+0.4$ \ci{0.0}{0.8} & 148/127 & $.2277$ \\
\addlinespace[2pt]
\multicolumn{5}{@{}l}{\textit{VIGA vs.\ One-Shot}} & \multicolumn{5}{@{}l}{\textit{VIGA vs.\ One-Shot}} \\
PSNR & 214 & $+3.02$ \ci{1.91}{4.21}$^\dagger$ & 147/67 & $4.69\!\times\!10^{-8}$ & PSNR & 238 & $-0.90$ \ci{-2.28}{0.33} & 111/127 & $.3309$ \\
SSIM & 214 & $+3.1$ \ci{1.8}{4.5}$^\dagger$ & 145/65 & $3.45\!\times\!10^{-8}$ & SSIM & 238 & $-3.2$ \ci{-5.4}{-1.1} & 105/133 & $.0799$ \\
Edge F1 & 214 & $+5.1$ \ci{2.8}{7.5}$^\dagger$ & 135/57 & $1.76\!\times\!10^{-8}$ & Edge F1 & 238 & $-3.2$ \ci{-5.1}{-1.4} & 110/128 & $.2704$ \\
LPIPS & 214 & $+32.0$ \ci{23.6}{40.4}$^\dagger$ & 152/62 & $6.61\!\times\!10^{-10}$ & LPIPS & 238 & $-9.9$ \ci{-22.0}{0.6} & 114/124 & $.5597$ \\
CLIP & 214 & $+0.4$ \ci{-0.2}{0.9} & 153/58 & $4.58\!\times\!10^{-11}$ & CLIP & 238 & $-0.5$ \ci{-0.8}{-0.2} & 109/129 & $.2180$ \\
\bottomrule
\end{tabular}}
\end{table}
 
\section{One-Shot editing strategy: statistics behind Section~\ref{sec:strategy}}
\label{app:strategy}
\paragraph{Measure.}
For this program-form analysis, each source line is truncated at its first \texttt{\#}, whitespace is normalized, and empty lines are removed; thus inline comments are stripped. \emph{Retention} counts initial-line occurrences whose normalized text appears anywhere in the generated program, divided by the initial line count. \emph{Changed lines} counts initial-line occurrences absent from the generated program plus generated-line occurrences absent from the initial program. Identical means equality of the normalized line sequences. A program is \emph{scene-clearing} if the source contains \texttt{read\_factory\_settings}, \texttt{bpy.ops.object.delete}, \texttt{objects.remove(}, or \texttt{nodes.clear()}. A program \emph{contains a 4.x binding} if it names a Blender identifier introduced in 4.0--4.2 and absent in 3.6 (\texttt{BLENDER\_EEVEE\_NEXT}, \texttt{surface\_render\_method}, the Principled-BSDF sockets \texttt{Specular IOR Level}, \texttt{Subsurface Weight}, \texttt{Subsurface Scale}, \texttt{Transmission Weight}, \texttt{Coat Weight}, \texttt{Coat Roughness}, \texttt{Sheen Weight}, \texttt{Emission Color}, and the Noise-Texture socket \texttt{Lacunarity}). The 100 geometry and material tasks are used because they are the tasks whose archived GPT One-Shot programs were produced and executed under Blender 3.6.
 
\begin{table}[tbp]
\centering\small
\caption{\textbf{One-Shot program form.} GPT-5.6-luna on the 100 geometry and material tasks under Blender 3.6, and the first-round program of the same backbone inside \vetr{} $N{=}1$ on the same tasks. Valid: tasks with a render for One-Shot; rendered programs / tasks run for round 1. First-error counts name the exception that ended the failed program.}
\label{tab:strategy}
\setlength{\tabcolsep}{5pt}
\resizebox{\ifdim\width>\linewidth\linewidth\else\width\fi}{!}{%
\begin{tabular}{l rr}
\toprule
 & One-Shot & \vetr{} $N{=}1$, round 1 \\
\midrule
Valid & 34 & 98 / 100 \\
Median retention of start lines & 0.10 & 0.96 \\
Programs retaining $\ge 50\%$ (valid) & 8 (7) & 99 \\
Programs retaining $< 50\%$ (valid) & 92 (27) & 1 \\
Programs identical to the start program & 0 & -- \\
Median changed lines & 188 & -- \\
Scene-clearing programs & 46 & -- \\
Programs containing a 4.x binding & 65 & 4 \\
\;\; \texttt{Specular IOR Level} / \texttt{Lacunarity} / \texttt{BLENDER\_EEVEE\_NEXT} & 44 / 13 / 12 & -- \\
First error = 4.x binding (same three) & 21 / 10 / 9 & -- \\
First error = other socket, type or syntax error & 18 & -- \\
Failed with no Python exception in the log (crash / kill) & 8 & -- \\
No-render generator rounds, all rounds of the 100 tasks & -- & 41 / 814 \\
\bottomrule
\end{tabular}}
\end{table}
 
\paragraph{Reading.}
The GPT One-Shot failures are concentrated in programs that discarded the start program: of the 92 programs retaining less than half of it, 65 fail, against 1 of the 8 that kept at least half. With \vetr{}'s prompt and loop, GPT produces first-round programs on all 100 tasks, with median retention 0.96 and 98 rendered outcomes; the full run has 41 logged no-render outcomes in 814 interactions on these tasks.
 
\section{Execution settings}
\label{app:env}
 
\begin{table}[tbp]
\centering\footnotesize
\caption{\textbf{Experiment map.} Each row identifies a separate source cohort. Iterative VIGA/\vetr{} runs allow ten generator interactions; $N$ is the number of candidate actions per interaction. Within each ablation, the backbone, Blender version, tools, prompts, and budget are fixed except for the named switch. Completion and paired task intersections are reported with the corresponding results.}
\label{tab:cohortmap}
\setlength{\tabcolsep}{3pt}
\begin{tabular}{@{}p{.19\linewidth}p{.13\linewidth}p{.15\linewidth}p{.43\linewidth}@{}}
\toprule
Evidence / source & Tasks & Blender & Configuration and comparison \\
\midrule
GPT systems (Table~\ref{tab:main}) & 280 each & 3.6 & GPT-5.6-luna; One-Shot, BlenderAlchemy, VIGA, \vetr{} $N=1/4$. Each system uses its own instructions and runner; BlenderAlchemy uses four candidates per interaction. \\[2pt]
Context ablations (Table~\ref{tab:component}) & 280 each & 3.6.0 & GPT, $N=1$; one default run and three variants (three-message generator history, shared generator/verifier memory, no visual verifier); quality scored after sampling finished. \\[2pt]
Scene origin (Table~\ref{tab:scene}) & 280 each & 3.6 & GPT, $N=1$; fixed base vs.\ saved scene; both retain 24 messages and use identical complete-program instructions. \\[2pt]
Guidance (Table~\ref{tab:guidance}) & 280 / 85 & 3.6 & GPT, $N=1$, fixed origin, 24 messages. VIGA role-prompt substitution on 280 tasks; recipe removal on 85 blend-shape tasks. Full is the scene-origin arm above. \\[2pt]
Qwen systems (Appendix~\ref{app:qwen}) & 280 each & 3.6 & Qwen3.8-27B; all five systems. VIGA/\vetr{} use thinking; BlenderAlchemy uses depth 2, breadth 2, one try, with thinking disabled. Timeout caps differ by system. \\
\bottomrule
\end{tabular}
\end{table}
 
\paragraph{Runtime configuration.}
All GPT and Qwen runs use Blender 3.6.0 with BlenderGym's Infinigen fork. BlenderGym's unmodified initial programs execute on all 280 tasks in the benchmark-compatible environment.
 
\paragraph{Result provenance.}
Each task/system has one designated source result. The GPT system table uses one complete Blender 3.6 run set, and the context ablations use one separately identified run set; failures in every source are retained. The GPT BlenderAlchemy row uses the same Blender 3.6 environment as the other current GPT systems, with four candidates per interaction. Per-task sources and scripts are enumerated in Appendix~\ref{app:artifacts}.
 
\paragraph{Answer extraction.}
\label{app:answer}
All iterative systems are scored using their last successful selected render from a completed run. A later execution failure contributes feedback and does not replace that answer. Tasks without a successful render, and timed-out or incomplete runs, are invalid. Stored renders are rescored by the same evaluator. Primary comparisons use the intersection of valid task identities; separate completion counts retain the information about missing answers.
 
\section{Within-run progress and guidance controls}
\label{app:guidance}
\paragraph{First executed program versus the final answer.}
We evaluate the first executed script in each Full trajectory from the scene-origin cohort. This proposal may follow an initial planning action but precedes feedback from any executed program in that run. It uses the same task, role instructions, backbone, and fixed-origin backend as the subsequent refinements. We score stored first-attempt renders irrespective of eventual completion: 258/280 first proposals render, 278/280 final runs are valid, and 256 tasks have both. No metric-based historical selection is applied.
 
\begin{table}[tbp]
\centering\small
\caption{\textbf{Refinement of the same initial proposals.} All 256 jointly available tasks; means are scaled by 100. Reduction compares final versus first executed program. Bootstrap, sign tests, and $\dagger$ use the same definitions as the main experiments. W/L/T favors the final answer.}
\label{tab:prefix}
\setlength{\tabcolsep}{4pt}
\resizebox{\ifdim\width>\linewidth\linewidth\else\width\fi}{!}{%
\begin{tabular}{@{}lrrcrr@{}}
\toprule
Metric & First program & Final answer & Reduction (\%) [CI] & W/L/T & $p$ \\
\midrule
$\nclip$ & 3.922 & 2.255 & $+42.5$ \ci{33.7}{50.6}$^\dagger$ & 190/50/16 & $2.11\!\times\!10^{-20}$ \\
$\PL$ & 1.710 & 0.829 & $+51.5$ \ci{38.4}{61.7}$^\dagger$ & 188/52/16 & $2.91\!\times\!10^{-19}$ \\
\bottomrule
\end{tabular}}
\end{table}
 
\paragraph{Role-instruction sensitivity.}
Two existing controls retain the Full runner, full generator history, $N=1$, and fixed execution origin. The first replaces the generator and verifier role-system prompts with the released VIGA role prompts (277/280 valid; 276 paired with Full). The second removes the body-skin recipe and worked example on the 85 blend-shape tasks (84 valid; 83 paired). Full completes 278/280 tasks overall. Table~\ref{tab:guidance} reports both primary metrics for both controls; neither control produces a significant paired difference. Source hashes and task-level configuration records accompany the measurements.
 
\begin{table}[tbp]
\centering\small
\caption{\textbf{Guidance controls under the same refinement configuration.} Positive reductions favor Full. Complete paired intervals and exact sign tests are shown, with ties excluded from the latter. The no-recipe control is restricted to blend-shape tasks.}
\label{tab:guidance}
\setlength{\tabcolsep}{4pt}
\resizebox{\ifdim\width>\linewidth\linewidth\else\width\fi}{!}{%
\begin{tabular}{@{}lrlcrr@{}}
\toprule
Control & $n$ & Metric & Full reduction (\%) [CI] & W/L/T & $p$ \\
\midrule
VIGA role prompts & 276 & $\nclip$ & $+9.4$ \ci{-5.4}{21.9} & 148/128/0 & $.2527$ \\
 & 276 & $\PL$ & $(6.7)$ \ci{-36.2}{14.8} & 145/130/1 & $.3986$ \\
Without body-skin recipe & 83 & $\nclip$ & $(0.7)$ \ci{-24.2}{19.4} & 42/39/2 & $.8243$ \\
 & 83 & $\PL$ & $+3.6$ \ci{-20.4}{24.7} & 43/40/0 & $.8264$ \\
\bottomrule
\end{tabular}}
\end{table}
 
\section{Identical-program replay of scene operations}
\label{app:replay}
We replay the archived first program of \texttt{geometry20}, which is byte-identical in the two closed-loop trajectories of Figure~\ref{fig:scene}(b). Its node transform sets scale $(0.5,1,1)$ and a $z$-rotation of $0.35$ radians, then calls \texttt{surface.add\_geomod}. The original base scene and script are held fixed. We execute it nine times under three policies: reload the base before each execution; retain the edited scene; or retain the scene but remove modifiers absent from the initial base before executing the unchanged script. This last policy supplies a compatible replacement contract. No language-model requests or rendering calls are made.
 
\begin{figure}[tbp]
\centering
\includegraphics[width=\linewidth]{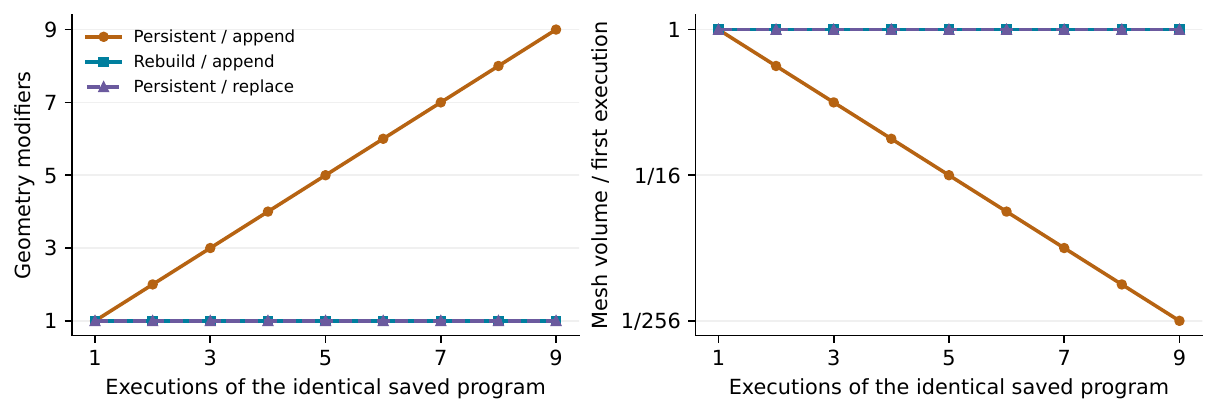}
\caption{\textbf{The same program has different repeated-execution semantics.} Measured modifier count and evaluated mesh volume on \texttt{geometry20}. Persistent appending creates one additional modifier per execution and halves the current volume each time. Rebuild and explicit modifier replacement coincide at one modifier and constant volume. All three use the same archived program and Blender 3.6.0; the volume is normalized by each policy's identical first execution.}
\label{fig:replay}
\end{figure}
 
Persistent execution yields one, three, and nine modifiers at steps 1, 3, and 9; the corresponding measured volumes are approximately 4, 1, and 0.015625 Blender units cubed. Rebuild and replacement both retain one modifier and volume 4 throughout. The evaluated meshes retain eight vertices and six faces under all policies: the effect here is repeated transformation, not growing mesh topology. Object bounds, modifier identities, script/base/helper hashes, and all 27 observations are supplied. This single-program intervention explains a concrete accumulation mechanism and identifies a compatible persistent alternative; the 280-task closed-loop experiment separately measures the overall backend policy.
 
\section{Artifacts and traceability}
\label{app:artifacts}
Table~\ref{tab:cohortmap} identifies the experiment associated with each result. The following records and scripts connect those results to their task-level sources.
\begin{myitemize}
\setlength{\itemsep}{2pt}
\item \textbf{GPT system metrics.} \path{supplement/gpt_single_runtime/metrics.json} contains all 1,400 task--system records in the Blender 3.6 comparison. The same directory contains per-task extended metrics, complete paired analyses, and saved cost counters. The 212-task intersection used for the $N{=}1$ comparisons is listed in \path{analysis.json}. The context-ablation and scene-origin experiments use their separately identified run sets.
\item \textbf{Paired tests.}
\path{paired_stats.py} produces task bootstraps and exact sign tests. Cost counters are in \path{run_stats.json}.
\item \textbf{Scene-origin ablation.} \path{supplement/scene_execution/metrics.json} contains all 560 task--arm records, including invalid tasks and hashes of scored renders. \path{paired_stats.py} in the same directory produces \path{paired_stats.json}, \path{paired_tasks.csv}, and both scene-ablation tables. The source run is \path{commit_gate_grid}, with arms \path{comp_full} and \path{comp_persistent_scene}; \path{source_audit/} preserves the runner, evaluator, task manifest, status records and representative logs. Figure~\ref{fig:scene} is built by \path{fig/scene_execution/build_figure.py} from stored images and candidate CSVs; its manifest records source hashes.
\item \textbf{Runtime audit.} The runtime validation records use
\path{env_check_start_scripts.py} outputs (\path{results.json} and renders) and \path{env_fix_report.py}.
\item \textbf{Extended metrics.}
\path{scripts/extended_metrics.py} scores stored final-answer images; per-task and paired records are under \path{extended_metrics/}.
Qwen sources are listed in Appendix~\ref{app:qwen}.
\item \textbf{Program-strategy examples.} Figure~\ref{fig:strategy} and Table~\ref{tab:strategy} use One-Shot programs under
\path{<run>/sampled_data/<task>/baseline/<model>.py}, iterative scripts under \path{<line>/<task>/scripts/<k>.py},
\path{generator_candidates.json}, and the corresponding One-Shot logs.
\end{myitemize}
 
\paragraph{Base scene and proposal contract.}
The task supplies both \texttt{blender\_file.blend} ($B$) and \texttt{start.py} ($p_0$). The latter is the editable reference program shown to the generator; the runner does not execute it as an extra preparation step before a proposal. It loads the supplied native scene and executes the submitted script once. Thus, a complete proposal is complete relative to the assets and objects already supplied in $B$. The answer rule in Appendix~\ref{app:answer} associates the evaluated render with the program in its selected delivered record, which may be a repaired program rather than the submitted script. A cached fallback retains the provenance of its earlier render and does not assign that render to a failed submission.
 
\paragraph{Pinned planning channel.}
The BlenderGym planning tool accepts a free-form \texttt{detailed\_plan} string covering visual analysis, code analysis, and intended edits. Its response is inserted into the second pinned message under \texttt{Initial plan:}; a subsequent planning call replaces that text in place. Both history settings use this mechanism.
 
\paragraph{Implementation-to-claim mapping.}
\texttt{agents/generator.py} samples and executes candidate actions, selects a returned outcome, invokes the verifier when requested, appends the selected interaction, and logs all candidates.
\texttt{agents/prompt\_builder.py} pins the initial two messages and applies the role-specific recent-message window, with tool-boundary handling.
The runner sets the generator window to three for the short-history variant, enables shared history for the role-asymmetry control, and bypasses verification for the no-verifier control.
The complete-script response field and latest-render guidance are prompt instructions; the final-answer evaluator separately selects the last successful render.
The $N{>}1$ path votes over action types before executing candidates of the selected type; Algorithm~\ref{alg:wide} explicitly places refinement before the selection of delivered outcomes. 
 
\paragraph{Prompt scope and system controls.}
The released generator prompts combine shared evidence-use instructions, Blender binding/API conventions, and family-specific guidance.
All context-window conditions share these instructions.
The VIGA runner saves successful scene updates, whereas the \vetr{} runner loads the task base scene for each complete-script execution.
History and scene-origin contrasts change individual settings; the same-run prefix analysis holds initial guidance fixed, and role-prompt/recipe controls are reported in Appendix~\ref{app:guidance}.
 
\end{document}